\pdfoutput=1
\documentclass[11pt]{article}

\usepackage[final]{acl}

\usepackage{times}
\usepackage{latexsym}

\usepackage[T1]{fontenc}
\usepackage[utf8]{inputenc}

\usepackage{microtype}

\usepackage{inconsolata}

\usepackage{graphicx}
\usepackage{amsmath}
\usepackage{mathtools}
\usepackage{booktabs}
\usepackage{enumitem}
\usepackage{hhline}
\usepackage{multirow}
\usepackage{bbm}
\usepackage{colortbl}
\usepackage{listings}
\usepackage{mdframed}
\usepackage{adjustbox}
\usepackage[most]{tcolorbox}

\definecolor{lightgray}{RGB}{240,240,240}
\definecolor{Gray}{gray}{0.7}
\definecolor{Green}{rgb}{0.05, 0.5, 0.06}
\definecolor{Purple}{rgb}{0.56, 0.0, 1.0}
\definecolor{Orange}{rgb}{1.0, 0.55, 0.0}\definecolor{Blue}{rgb}{0.0, 0.2, 0.4}

\title{\textsc{BottleHumor}: Self-Informed Humor Explanation using the Information Bottleneck Principle}
\author{EunJeong Hwang$^{1,2}$, Peter West$^{1}$, and Vered Shwartz$^{1,2}$ \\
$^1$ University of British Columbia~~~$^2$ Vector Institute for AI\\
{\tt \{ejhwang,pwest,vshwartz\}@cs.ubc.ca}}

\newcommand{\method}{\textsc{BottleHumor}}
\newcommand{\base}{\textsc{ZS}}
\newcommand{\chain}{\textsc{CoT}}
\newcommand{\critic}{\textsc{SR}}
\newcommand{\nocritic}{\textsc{SR-noC}}

\begin{document}
\maketitle
\begin{abstract}
% Humor plays a crucial role in understanding ourselves, society, and the world, shaped by cultural, social, and personal factors, and expressed through images, text, and audio. 
Humor is prevalent in online communications and it often relies on more than one modality (e.g., cartoons and memes).  
Interpreting humor in multimodal settings requires drawing on diverse types of knowledge, including metaphorical, sociocultural, and commonsense knowledge. However, identifying the most useful knowledge remains an open question. 
We introduce \method{}, a method inspired by the information bottleneck principle that elicits relevant world knowledge from vision and language models which is iteratively refined for generating an explanation of the humor in an unsupervised manner. Our experiments on three datasets confirm the advantage of our method over a range of baselines. 
Our method can further be adapted in the future for additional tasks that can benefit from eliciting and conditioning on relevant world knowledge and open new research avenues in this direction.

%We provide a detailed analysis of the IB components and the impact of implications on final explanations.

\end{abstract}

\section{Introduction}
\label{sec:intro}
% \textbf{Problem}:
% \paragraph{Novelty:}
% \paragraph{Related Work:}
% \paragraph{Contribution:}

Humor is an effective communication tool \cite{stauffer-humor, wanzer-humor, vartabedian-humor, kasulis-humor} that can manifest in various forms, including puns, exaggerated facial expressions, absurd behaviors, and incongruities \cite{philosophy-humor}. It is shaped by multiple factors such as culture, social interactions, societal phenomena, and personal imagination \cite{humorous-things, Warren-humor}. %, and can be expressed in images and text both explicitly and implicitly. Because of these factors, its interpretation varies based on age, cultural background, and context. While individuals may perceive humor differently, it remains an effective communication tool when used appropriately \cite{stauffer-humor, wanzer-humor, vartabedian-humor, kasulis-humor}.

\begin{figure}[t]
  \includegraphics[width=\linewidth]{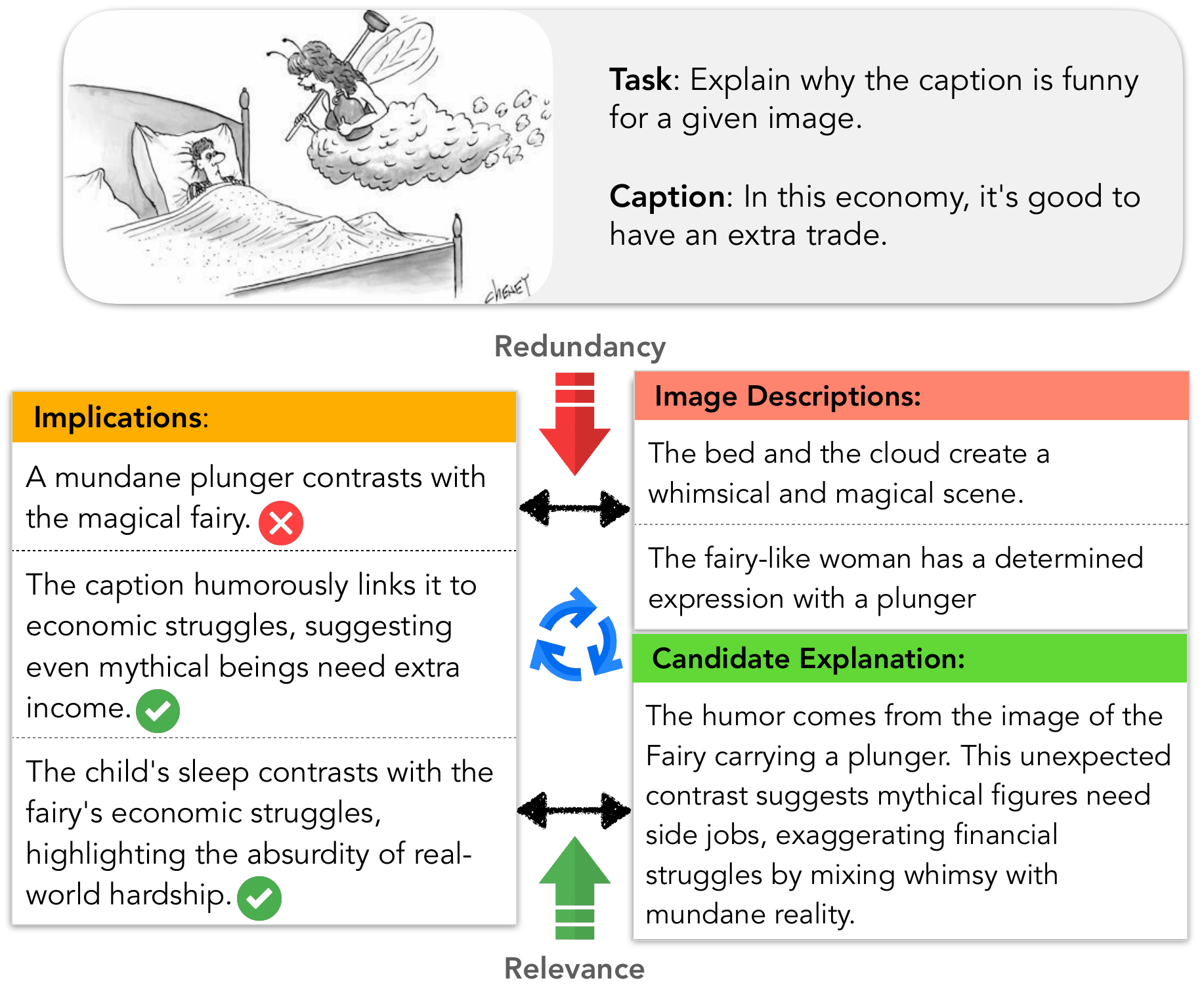} \hfill
  % \vspace{-20pt}
  \caption{Humor understanding requires understanding world knowledge. \method{} aims to reduce redundancy in existing inputs (e.g. image descriptions) while increasing relevance to candidate explanations.}
  \label{fig:fig1}
\end{figure}

In particular, humor is prevalent in online communications \cite{mcculloch2020because}, often spanning multiple modalities \cite[e.g., cartoons and memes;][]{shifman2013memes}. Interpreting humor across modalities requires ``reading between the lines'', connecting textual and visual elements to grasp the meaning \cite{Warren-humor}. 
For example, in Fig.~\ref{fig:fig1}, connecting the tooth fairy depicted in the image carrying a plunger to the caption, ``In this economy, it's good to have an extra trade'', creates the humorous interpretation that in this state of the economy, even the imaginary fairy needs a side job as a plumber.
% For example, in Fig.~\ref{fig:fig1}, the humor arises from recognizing the tooth fairy carrying a plunger in the image and linking it to the caption, ``In this economy, it's good to have an extra trade'', meaning even the imaginary fairy needs a side job as a plumber due to economic hardship.

Several datasets for multimodal humor understanding tasks were proposed, where models are tasked with generating free-text humor explanations for an image and a caption \cite{hwang-shwartz-2023-memecap, hessel-etal-2023-androids, nandy-etal-2024-yesbut, hu2024cracking}. However, they are often overlooked in vision-and-language models (VLMs) evaluations, possibly due to the subjective nature of humor and the challenges in evaluating free-text explanations. With that said, VLMs have demonstrated remarkable visual reasoning capabilities on datasets requiring scientific knowledge \cite{scienceqa}, commonsense knowledge \cite{okvqa}, and spatial reasoning \cite{Liu2022VisualSR} and there is a prominent line of work on enhancing multimodal reasoning \cite{zhang2024multimodal, MitraCCoT, kam-cot, hu2024visual}.

In this paper, we introduce \method{}, a method inspired by the information bottleneck (IB) principle. \method{} leverages VLMs to generate and iteratively refine implications and explanations from an image and text, selecting those most relevant for explaining the humor in the image and maximizing information gain.
% These implications iteratively refine the generated explanations and vice versa.   
As an off-the-shelf method, it is applicable to any VLM. 

We evaluate \method{} on three multimodal humor explanation datasets: MemeCap \cite{hwang-shwartz-2023-memecap}, NewYorker \cite{hessel-etal-2023-androids}, and YesBut \cite{nandy-etal-2024-yesbut}. Prior work relied on reference-based automatic metrics that overlook lexical variability and the open-endedness of explanations and costly human evaluation. Leveraging the strong text understanding capabilities of LLMs, we propose new automatic evaluation metrics that resemble precision and recall, and better correlate with human judgments. \method{} improves $F_1$ by up to 8.2, 4.3, and 2.8 points on MemeCap, NewYorker, and YesBut, respectively, compared to zero-shot baselines and outperforms existing self-refine methods that merely iterate on and refine the explanation without generating intermediate implications. Our results highlight the importance of incorporating implications, paving the way for future research on incorporating diverse world knowledge in complex reasoning tasks.\footnote{Our code and data are available at:\\ \url{https://github.com/eujhwang/bottle-humor}}
% \footnote{We will make our code and data available upon publication.}
% Uncomment for the camera-ready version

% Our method lays the foundation for future research exploring the discovery of diverse world knowledge in complex problems requiring multifaceted knowledge integration.

% (3.4+4+1.4+8.2)/4 = 4.2
% (4.3+0.1+0.6)/3 = 1.7
% (2.4+2.0+2.8+1.4)/4 = 2.15
% reference : https://aclanthology.org/2024.findings-naacl.73.pdf
% 

\section{Related Work}
\label{sec:bg}
\paragraph{Multimodal Humor Understanding.} Earlier works on humor understanding primarily focus on detection in images and videos \cite{Chandrasekaran_2016_CVPR, castro-etal-2019-towards, patro-humor}. 
Recent work shifted to generative tasks, typically explaining humor in an image \cite{hwang-shwartz-2023-memecap, hessel-etal-2023-androids, nandy-etal-2024-yesbut} or video \cite{smile, hasan-etal-2019-ur}. Understanding and explanation generation remain underexplored due to the complexity of the task and free-text evaluation. The V-Flute dataset \cite{vflute} addresses this by re-casting this as predicting whether an image containing humorous elements or visual metaphors \emph{entails} a given description, while providing justification. We focus on the generative version of this task, proposing a method to enhance humor explanation and a framework for automatic evaluation.

% \citet{li-etal-2024-enhancing-advanced} introduce multimodal in-context learning but focus on discriminative tasks such as Winoground \cite{winoground} and Visual Commonsense Reasoning \cite{vcr}.

\paragraph{Iterative LLM-based Reasoning.} Many methods elicit knowledge from the LLM for intermediate reasoning steps.  
\citet{shwartz-etal-2020-unsupervised} elicited clarification questions and answers, then incorporated these in the input. Modern Few-shot prompting removed the need for supervision for these explanations \cite{marasovic-etal-2022-shot,wiegreffe-etal-2022-reframing}. One popular approach is Chain-of-Thought \cite[CoT;][]{cot}. CoT steers LLMs to generate intermediate reasoning steps towards the final answer, improving multi-step arithmetic, commonsense, and symbolic reasoning tasks. Relevant successor approaches include self-refine \cite{madaan2023selfrefine} which prompts LLMs to iteratively improve their answers with self-generated feedback. Eliciting knowledge from LLMs to improve predictions has been used for opinion understanding \cite{hwang2024graph, hoyle-etal-2023-natural}, factuality \cite{akyurek-etal-2024-deductive}, and consistency \cite{liang-et-al-consistency}.

CoT has been adapted to the vision and language setting \cite{zhang2024multimodal} by adding external knowledge \cite{kam-cot}, extracting a scene graph \cite{MitraCCoT}, or using visual sketches as intermediate reasoning steps \cite{hu2024visual}. 
Most existing works focus on benchmarks such as ScienceQA \cite{scienceqa} and visual commonsense reasoning \cite{okvqa}, with (a) definitive/objective answers; and (b) simple evaluation metrics (e.g., ScienceQA is multiple-choice). We focus on multimodal explanation generation tasks in which the answers are open-ended and nuanced. As in CoT, we elicit intermediate reasoning steps from the models, but propose a novel method using the information bottleneck principle to guide generation and selection of useful knowledge for a correct explanation. 

\paragraph{Information Bottleneck Principle.} The Information Bottleneck principle \cite[IB;][]{ib}, based on information theory, extracts relevant information from an input while minimizing redundancy (Sec.~\ref{sec:ib}). It has been applied to a wide range of tasks \cite{ib-review}, including representation learning \cite{graph-ib, lee2021compressive}, deep learning \cite{michael2018on, icml2023kzxinfodl}, summarization \cite{west-etal-2019-bottlesum, ju-etal-2021-leveraging-information, li-etal-2021-ease}, speech recognition \cite{hecht09c_interspeech}, and multimodal learning \cite{Mai_2023, FangWZHZXW024}. Most prior works apply the IB principle during training to learn useful feature representations, with the exception of \citet{west-etal-2019-bottlesum, ju-etal-2021-leveraging-information}, who use IB for unsupervised summarization. In this work, we extend the IB principle to multimodal humor understanding to identify relevant LLM world knowledge.

\section{\method{}}
\label{sec:method}
\begin{figure*}[t]
  \includegraphics[width=\linewidth]{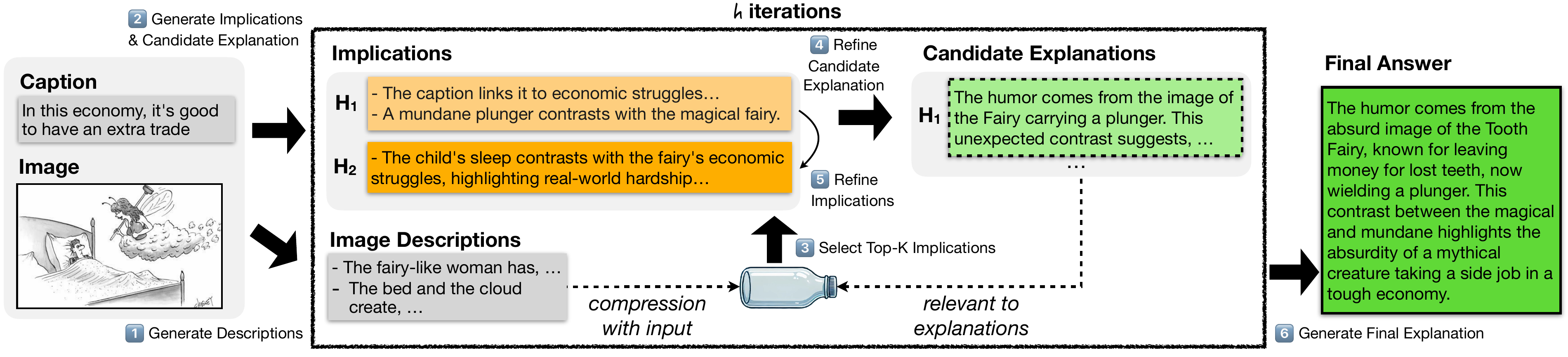} \hfill
  \vspace{-8pt}
  \caption {Overview of \method{}. We begin by generating descriptions, implications, and a candidate explanation (steps 1 and 2). Then, we refine the implications and candidate explanations over $h$ iterations using the IB principle (steps 3 to 5), ultimately generating a final explanation from the refined implications and candidate explanations (step 6).}
  \label{fig:overview}
\end{figure*}

Given a humoristic image along with an accompanying text (\textit{caption}), our goal is to generate a descriptive explanation of the humor. For example, in Figure~\ref{fig:overview}, a fairy woman with a plunger looking at a boy can be humorously explained as ``The humor comes from a fairy with a plunger, taking a side job because of a tough economy'' \cite[from the NewYorker dataset;][]{hessel-etal-2023-androids}.

We propose \method{} (Figure~\ref{fig:overview}), a multihop reasoning method inspired by the IB principle (Sec.~\ref{sec:ib}).
We integrate the visual and textual components to generate implications (Sec.~\ref{sec:method:gen_imps}). We then select the most useful implications by employing the IB principle (Sec.~\ref{sec:method:refine-imps}), and add them to the input to generate candidate explanations  (Sec.~\ref{sec:method:gen_response}). This iterative process alternates between refining implications and explanations.

\subsection{The Information Bottleneck Principle}
\label{sec:ib}

We use the Information Bottleneck principle \cite[IB;][]{ib} to select useful implications in \method{}.
IB aims to extract the most relevant information from a given input variable while minimizing redundancy. Specifically, IB seeks to compress the input source $S$ into a representation $\hat{S}$ while retaining the information most relevant to predicting the target $Y$. This objective is formulated as minimizing the following equation:
% \vspace{-1mm}
\begin{align}
    \notag
    I(S, \hat{S}) - \alpha I(\hat{S}, Y)
\end{align}
where $I$ denotes mutual information, and $\alpha$ is a parameter to balance compression term $I(S, \hat{S})$ with relevance term $I(\hat{S}, Y)$.

% \paragraph{Notation}
% Let $M$ represent the total number of images. For each image $I_m$, where $m$ ranges from 1 to $M$, the following components are associated:

% - The \textbf{caption} of the image is denoted as $C_m$.

% - The \textbf{description} of the image is represented as $D_m$, which consists of sentences $\{ d_m^1, d_m^2, \dots, d_m^{i} \}$. Here, $i$ is the total number of sentences in the description.

% - The \textbf{implications} at hop $h$ are denoted as $P_m^{h,j} = \{ p_m^{h,1}, p_m^{h,2}, \dots, p_m^{h,j} \}$, where $p_m^{h,j}$ represents the $j$-th implication at hop $h$.

% - The \textbf{candidate explanation} set at hop $h$ is represented as $R_m^{h,k} = \{ r_m^{h,1}, r_m^{h,2}, \dots, r_m^{h,k} \}$. Here, $r_m^{h,k}$ is the $k$-th candidate explanation at hop $h$.

\subsection{Eliciting Multi-Hop Implications} 
\label{sec:method:gen_imps}

First, we generate a set of natural language implications of the input.
The goal of this step is to discover connections across different objects, concepts, and situations described in the input.

\paragraph{Image Descriptions.} 
As a first step, we provide the image $I$ to a VLM to generate a detailed \textit{image description} $D$, focusing on the scene and objects while ignoring the humoristic meaning behind the image. We limit the description to a maximum of five sentences. 

\paragraph{Implications.} 
Using these descriptions, the VLM elicits \textit{implications}: commonsense knowledge, social norms, and possible connections for the objects in the description $D$ and the caption $C$. Implications generated at hop $h$ are denoted as $P^{h} = \{ p_1^{h}, p_2^{h}, \dots, p_j^{h} \}$.

In the first hop, the implications are derived from the image $I$, its caption $C$, and a subset of two image descriptions $D$, selected via a sliding window to balance efficiency (i.e., input length and cost) and coverage. 
From the second hop onward, we provide the VLM with \textit{candidate explanations} (see below) and one of the previously selected top-$k$ implications (Sec.~\ref{sec:method:refine-imps}). 

When the number of generated implications exceeds 15, we cluster them using sentence embeddings and select the implications closest to each cluster's centroid. This step reduces redundancy while preserving diversity.

%commonsense knowledge, or social norms associated with different objects, concepts, or situations, referred to as \textit{implications}. 

\paragraph{Candidate Explanations.} To guide implication selection for generating the correct output, we provide the image $I$ and caption $C$ to the VLM to generate a set of \textit{candidate explanations} at each hop: $R^{h} = \{ r_1^{h}, r_2^{h}, \dots, r_k^{h} \}$. One candidate explanation acts as an initial hypothesis, refined iteratively when additional information (implications) becomes available.
In the first hop, we generate candidate explanations by providing the VLM with the image $I$, caption $C$, and descriptions $D$. From the second hop onward, we condition---in addition to the previous inputs---on each of the $k$ implications selected in the previous hop (\S\ref{sec:method:refine-imps}) to generate $k$ candidate explanations. 
The prompts used for generating image descriptions, implications, and candidate explanations are in Appendix \ref{app:gen-prompts}.

\subsection{Selecting and Refining Useful Implications}
\label{sec:method:refine-imps}

We aim to select the top $k$ most useful implications at each hop, which should add meaningful information beyond the image and caption while providing relevant context for generating a target response. These requirements lend themselves to the two core IB components: compression and relevance.

\paragraph{Compression.} 

The compression term is used to ensure that new implications provide additional information beyond what is already known. We measure the redundancy of each implication generated in the current hop $h$, $\{P_j^{h}\}_{j=1}^J$ with the inputs $X^h = \{C, D, P^{h-1}\}$, which include the image, caption, and implications generated at previous hops (when applicable). We can think of this as testing whether the new set $X^h + \{P_j^{h}\}_{j=1}^J$ can be easily compressed back to $X^h$ (redundant). To that end, we embed each of the inputs using sentence embeddings and compute the maximum cosine similarity between the target implication and each input in $X^h$, representing the maximum redundancy with existing information:
% \[
% I(X, Z) = \min(\max_{i\in I, j \in J}(\text{CosSim}(X_{i}, P_{j}^{h})))
% \]
\[
\hat{I}(X, P_{j}^{h}) = \max_{i\in I}(\operatorname{cos}(X_{i}, P_{j}^{h}))
\]

\paragraph{Relevance.} The relevance term is used to ensure that implications provide useful information for generating a target explanation. Since our method is unsupervised, we use the VLM to generate candidate explanations at hop $h-1$: $Y = \{R^{h-1}\}_{i=1}^I$, which we use as a proxy for the gold standard answer in the next hop $h$. We measure the relevance of the target implication $P_j^{h}$ as the maximum probability (minimum cross entropy loss) for predicting the candidate explanation from the current (textual) inputs 
$\hat{Z}_j^{h} = \{C, D, P^{h-1}, P_j^{h}\}$, which include the caption, image description,  implications from previous hops, and the target 
implication: 
% \[
% I(Z, Y) = \min(\text{CE}(R_m^{h-1} \mid \sigma(\hat{Z}))
% \]
\[
\hat{I}(P_{j}^{h}, Y) = \min_{i \in I} (\operatorname{CE}(R_i^{h-1} \mid \hat{Z}_j^{h}))
\]
Cross-entropy values tend to be lower for short candidate explanations, leading to abnormally low scores for low-quality responses. To address this, we introduce a length penalty to adjust for deviations from the average response length. Responses significantly shorter or longer than the average receive a larger penalty.
We incorporate a scaling factor $\beta$, defined as the ratio of the average cross-entropy to the average length. The length penalty is then formulated as:
\[
LP_i = \beta \cdot |L_i - \bar{L}|, \quad \beta = \frac{\bar{CE}}{\bar{L}}
\]
\noindent where $L_i$ is a length for $i$-th candidate explanation, $\bar{L}$ is the mean token length across all candidate explanations, and $\bar{CE}$ is the mean cross-entropy loss across all candidate explanations.
The final relevance term for each implication becomes:
\[
\hat{I}(P_{j}^{h}, Y) = \min_{i \in I} (\operatorname{CE}(R_i^{h-1} \mid \hat{Z}_j^{h}) + LP)
\]
% \[
% I(Z, Y) = \min(\text{CE}(R_m^{h-1} \mid \sigma(\hat{Z})) + LP)
% \]
We use the open/efficient Qwen2-1.5B \cite{yang2024qwen2technicalreport} LLM to compute cross-entropy values.

\paragraph{Selecting Implications.} With these compression and relevance terms, we formulate the final IB-based objective function. Since the goal is to minimize redundancy (maximize compression) and maximize relevance, we select $k$ implications based on the following equation:
\begin{align}
\label{eq:select_impl}
& {\min\limits_{\underset{}{k}}}_{j \in J} \hat{I}(X, P_{j}^{h}) - \hat{I}(P_{j}^{h}, Y) = \\  
& {\min\limits_{\underset{}{k}}}_{j \in J} \left\{
\begin{array}{l}
    \max_{i\in I}(\operatorname{cos}(X_{i}, P_{j}^{h}))~+ \\
    \alpha \min_{i \in I} (\operatorname{CE}(R_i^{h-1} \mid \hat{Z}_j^{h}) + LP)
\end{array}
\right\} \notag
% \label{eq:select_impl}
\end{align}
\noindent where $\alpha$ is a hyperparameter that controls the trade-off between the compression and relevance terms. In our experiments, we set $\alpha = 0.7$, based on our empirical observation. A detailed analysis of the effect of varying $\alpha$ is provided in Appendix~\ref{app:alpha-effect}.

We use the implications in each hop to refine the candidate explanations in the next hop and vice versa. To avoid excessive calculation during the implication refinement step, we keep the number of candidate explanations to a maximum of three based on the cross entropy scores computed using all existing inputs. These inputs, denoted as $\hat{Z}_j^{h} = \{C, D, P_j^{h}, R_i^{h-1}\}$, include caption, image descriptions, current hop implications, and previous hop candidate explanations. We then select top-$k$ candidate explanations ($k=3$) in current hop candidate explanations $R_i^{h}$ that minimize the cross-entropy: 
\begin{equation}
R_{\text{top-}k}^{h} = \arg min_{i \in I, |I| = k} \operatorname{CE}(R_i^{h} \mid \hat{Z}_j^{h})
\label{eq:select_response}
\end{equation}

\noindent In our experiments, we set the number of hops $H$ to 2 and the number of reasoning chains $k$ to 3. %, meaning that we find top-3 paths of 2-hop implications.  we set the number of reasoning hops $H$ to 2 and the number of reasoning chains $k$ to 3

\subsection{Generating Final Answer}
\label{sec:method:gen_response}

After $H$ iterations of refinement, we generate the final answer. As for candidate explanation generation in earlier hops, we provide the VLM with the image $I$, its caption $C$, the $k$ implications selected in the previous hop (Eq.~\ref{eq:select_impl}), and the $k$ candidate answers selected in the previous hop (Eq.~\ref{eq:select_response}), instructing it to generate a response.

We used Sentence Transformer\footnote{BAAI/bge-large-en-v1.5} for all sentence embeddings. The prompts for generating multi-hop implications and explanations are in Appendix \ref{app:gen-prompts}.

\section{Experimental Setup}
\label{sec:exp_setup}
\subsection{Datasets}
\label{sec:exp_setup:datasets}

We evaluate \method{} on three multimodal humor datasets (see examples in Appendix \ref{app:dataset-eg}):

\paragraph{MemeCap \cite{hwang-shwartz-2023-memecap}.} Each instance includes a meme paired with a title (social media post to which the meme was attached). The task is to generate a brief explanation, compared against multiple reference explanations. The task requires interpreting visual metaphors in relation to the text, where models can benefit from reasoning about background knowledge. 
% , and reference captions in the test images can contain up to four captions. 
% includes 5.8k training and validation samples, along with 559 test samples. 

\paragraph{New Yorker Cartoon \cite{hessel-etal-2023-androids}.} We focus on the explanation generation task: given a New Yorker cartoon and its caption, generate an explanation for why the caption is funny given the cartoon, requiring an understanding of the scene, caption, and commonsense and world knowledge.  
%indirect and playful meanings tied to human experience and culture.
% It includes 2.3k training samples, 130 validation samples, and 130 test samples, and offers five different splits. 

\paragraph{YesBut \cite{nandy-etal-2024-yesbut}.} 
%dataset consists of 1k satirical and 1k non-satirical test samples, with our focus on the satirical test set. 
Each instance contains an image with two parts captioned ``yes'' and ``but''. The task is to explain why the image is funny or satirical.
% requiring an understanding of commonsense knowledge, social norms, and cultural references related to everyday objects and situations.

Since our method is unsupervised, we use the test set portions of these datasets. Due to resource and cost constraints, we don't evaluate our method on the full test sets. Instead, from each dataset, we randomly sample 100 test instances. We repeat the process three times using different random seeds to obtain three test splits and report average performance and standard deviation.

\subsection{Models}
\label{sec:exp_setup_models}

We test our method with two closed-source and two open-source VLMs.
\paragraph{GPT-4o \cite{hurst2024gpt}} is an advanced, closed-source multimodal model processing text, audio, images, and video and generating text, audio, and images. It matches GPT-4's performance in English text tasks with improved vision understanding.
\paragraph{Gemini \cite{team2023gemini}} is a closed-source multimodal model from Google, available in multiple variants optimized for different tasks. 
We use \texttt{Gemini 1.5 Flash} for evaluation and \texttt{Gemini 1.5 Flash-8B} for experiments, a smaller, faster variant with comparable performance.
\paragraph{Qwen2 \cite{yang2024qwen2technicalreport}} is an open-source multimodal model built on a vision transformer with strong visual reasoning. We use the \texttt{Qwen2-VL-7B-Instruct} model, competitive with GPT-4o on several benchmarks.
\paragraph{Phi \cite{phi}} is a lightweight, open-source 4.2B-parameter multimodal model, trained on synthetic and web data. We use \texttt{Phi-3.5-Vision-Instruct}, optimized for precise instruction adherence.

\subsection{Baselines}
\label{sec:exp_setup:baselines}

We compare our method to four prompting-based baselines:\footnote{Temperature set to 0.8 for all baselines.} zero-shot (\base{}), Chain-of-Thought (\chain{}) prompting, and self-refinement with (\critic{}) and without (\nocritic{}) a critic.

\base{} generates a final explanation directly from the image and caption using VLM. \chain{} follows a similar setup but instructs the model to produce intermediate reasoning chains \cite{cot}. 
Additionally, we implement \critic{}, a multimodal variant of self-refinement \cite{madaan2023selfrefine}, where a \textit{generator} produces a response, and a \textit{critic} evaluates it based on predefined criteria. The critic's feedback helps refine the output iteratively\footnote{Refinement steps set to 2 for fair comparison.}. Evaluation criteria include correctness, soundness, completeness, faithfulness, and clarity (details in Appendix \ref{app:base-prompts}).
\nocritic{} functions identically to \critic{} but without a \textit{critic model}, refining candidate explanations without feedback. This also serves as an ablation of the implications from \method{}. Prompts for baselines are in Appendix \ref{app:base-prompts}.

\subsection{Evaluation Metrics}
\label{sec:exp_setup:eval}
While human evaluation is often the most reliable option for open-ended tasks like ours \cite{hwang-shwartz-2023-memecap}, it is costly at scale. LLM-based evaluations (e.g., with \texttt{Gemini 1.5 Flash}) offer a more affordable alternative but are not always reliable \cite{biases_paper}. Prior research in fact verification has found that modern closed-source LLMs excel at fact checking when the complex facts are decomposed into simpler, atomic facts and verified individually \cite{gunjal-durrett-2024-molecular, samir-etal-2024-locating}. Inspired by this approach, we propose LLM-based precision and recall scores.

For recall, we decompose the reference $ref$ into atomic facts: $\{y_1, y_2, ..., y_n\}$ and check whether each appears in the predicted response $pred$.
% The percent of facts present in the predicted response forms the recall score:
\[
\text{Recall} = \frac{1}{n} \sum_{i=1}^{n} \mathbbm{1} \big( LLM(y_i, pred) = \text{Yes} \big)
\]
where $n$ is the number of atomic facts in $ref$.

Precision follows the same process in reverse, decomposing $pred$ into a list of atomic facts: $\{x_1, x_2, ..., x_m\}$ and verifying their presence in $ref$:
\[
\text{Precision} = \frac{1}{m} \sum_{i=1}^{m} \mathbbm{1} \big( LLM(x_i, ref) = \text{Yes} \big)
\]
where $m$ is the number of atomic facts in $pred$. Both decomposition and verification use Gemini-Flash-1.5 with a temperature of 0.2.

In preliminary experiments, we observed that human references tend to omit obvious visual details, whereas model-generated answers are often more complete, referencing visual information. To prevent penalizing the models for these facts, we incorporate literal image descriptions (Sec~\ref{sec:method}) into the reference by decomposing them and adding them to the atomic facts for fairer evaluation. Based on the precision and recall scores, we report the macro-$F_1$ score.

To assess the reliability of our metrics, we conducted a human evaluation on 130 random samples across all models and datasets via CloudResearch (details in Appendix \ref{app:cloudresearch}). Human annotators determined whether each atomic sentence appeared in the corresponding text (e.g., reference). The average agreement between the LLM-based evaluator and two human annotators was 77.1\% (\(\kappa = 54.1\)), similar to the agreement between the two annotators: 75.4\% (\(\kappa = 50.8\)), indicating considerable alignment with human judgment. Prompts are in Appendix \ref{app:eval-prompts}.

\section{Results}
\label{sec:results}

\begin{table*}[!t]
    \small
    \centering
    \setlength{\tabcolsep}{4pt}
    \begin{tabular}{ll|cc>{\columncolor[gray]{0.9}}c|cc>{\columncolor[gray]{0.9}}c|cc>{\columncolor[gray]{0.9}}c|c}
    \toprule
    
   &   & \multicolumn{3}{c}{\textbf{MemeCap}} &  \multicolumn{3}{|c|}{\textbf{NewYorker}} &  \multicolumn{3}{c|}{\textbf{YesBut}} & \textbf{Avg.} \\  
   \rowcolor{white} \textbf{Model} &  \textbf{Method} & \textbf{P} & \textbf{R} & $\mathbf{F_1}$ &  \textbf{P} & \textbf{R} & $\mathbf{F_1}$ & \textbf{P} & \textbf{R} & $\mathbf{F_1}$ & $\mathbf{F_1}$ \\ \midrule
\multirow{5}{*}{\textbf{GPT4o}} & 
ZS  &
81.8$_{2.0}$ & 34.1$_{2.6}$ & 48.1$_{2.8}$ & 
75.4$_{1.4}$ & 42.0$_{2.5}$ & 53.9$_{1.9}$ & 
73.9$_{2.3}$ & 47.8$_{6.6}$ & 58.0$_{5.5}$ & 53.3$_\text{1.9}$ \\
 
& CoT  & 
78.6$_{2.1}$ & 34.1$_{1.1}$ & 47.5$_{1.1}$ & 
\textbf{76.0}$_{0.7}$ & 26.1$_{2.7}$ & 38.8$_{3.1}$ & 
\textbf{74.1}$_{0.9}$ & 26.2$_{2.9}$ & 38.7$_{3.2}$ & 41.7$_\text{1.2}$ \\

& SR  & 
75.3$_{1.3}$ & 31.8$_{0.9}$ & 44.8$_{0.9}$ & 
75.1$_{1.2}$ & 44.6$_{1.7}$ & 56.0$_{1.7}$ & 
69.4$_{1.3}$ & 46.7$_{3.9}$ & 55.8$_{3.1}$ & 52.2$_\text{1.0}$\\

& SR-noC  & 
\textbf{81.9}$_{2.6}$ & 33.5$_{0.5}$ & 47.5$_{0.7}$ & 
75.2$_{1.2}$ & 45.8$_{1.0}$ & 56.9$_{0.5}$ & 
72.4$_{1.6}$ & 46.3$_{2.6}$ & 56.5$_{2.4}$ & 53.6$_\text{1.1}$\\

& \method{}  & 
79.1$_{2.6}$ & \textbf{38.2}$_{0.8}$ & \textbf{51.5}$_{0.3}$ & 
74.5$_{2.2}$ & \textbf{47.7}$_{0.3}$ & \textbf{58.2}$_{0.5}$ & 
73.8$_{3.5}$ & \textbf{51.2}$_{2.9}$ & \textbf{60.4}$_{2.6}$ & \textbf{56.7}$_\text{1.1}$\\ \midrule

\multirow{5}{*}{\textbf{Flash1.5}} & 
ZS  & 
79.2$_{1.7}$ & 17.7$_{2.3}$ & 28.9$_{3.0}$ & 
76.6$_{1.2}$ & \textbf{24.1}$_{3.1}$ & 36.6$_{3.7}$ & 
74.5$_{1.8}$ & 28.5$_{3.5}$ & 41.1$_{3.7}$ & 35.5$_\text{0.4}$\\

& CoT  & 
79.5$_{1.8}$ & 16.1$_{1.0}$ & 26.7$_{1.5}$ & 
\textbf{76.7}$_{2.6}$ & 13.1$_{0.5}$ & 22.3$_{0.8}$ & 
\textbf{77.3}$_{2.6}$ & 16.4$_{1.5}$ & 27.1$_{2.2}$ & 25.4$_\text{0.7}$\\

& SR  & 
76.2$_{2.1}$ & 19.4$_{1.2}$ & 30.9$_{1.6}$ & 
73.7$_{1.7}$ & 22.9$_{1.5}$ & 34.9$_{1.8}$ & 
72.7$_{0.9}$ & 28.9$_{3.5}$ & 41.3$_{3.5}$ & 35.7$_\text{1.1}$\\

& SR-noC  & 
\textbf{80.9}$_{0.7}$ & 19.4$_{0.5}$ & 31.3$_{0.7}$ & 
74.0$_{2.0}$ & 21.2$_{1.1}$ & 32.9$_{1.5}$ & 
71.3$_{1.1}$ & 26.5$_{4.7}$ & 38.5$_{5.0}$ & 34.3$_\text{2.3}$\\

& \method{}  & 
79.6$_{0.7}$ & \textbf{20.8}$_{1.8}$ & \textbf{32.9}$_{2.2}$ & 
76.2$_{1.0}$ & \textbf{24.1}$_{1.0}$ & \textbf{36.7}$_{1.3}$ & 
73.4$_{1.8}$ & \textbf{30.6}$_{4.6}$ & \textbf{43.1}$_{4.7}$ & \textbf{37.6}$_\text{2.7}$\\ \midrule

\multirow{5}{*}{\textbf{Qwen2}} & 
ZS  & 
74.3$_{2.1}$ & 22.8$_{1.9}$ & 34.8$_{2.4}$ & 
66.7$_{1.0}$ & \textbf{19.4}$_{0.2}$ & \textbf{30.1}$_{0.3}$ & 
70.0$_{1.8}$ & 19.7$_{0.5}$ & 30.7$_{0.4}$ & 31.9$_\text{1.2}$\\

& CoT  & 
71.6$_{4.0}$ & 22.0$_{1.2}$ & 33.6$_{1.5}$ & 
\textbf{70.9}$_{1.5}$ & 11.0$_{1.4}$ & 19.0$_{2.1}$ & 
\textbf{72.2}$_{1.6}$ & 13.6$_{4.4}$ & 22.7$_{6.3}$ & 25.1$_\text{2.6}$\\

& SR  & 
73.1$_{1.5}$ & 23.9$_{1.2}$ & 36.1$_{1.5}$ & 
67.4$_{0.9}$ & 17.8$_{1.1}$ & 28.2$_{1.4}$ & 
68.9$_{0.7}$ & 20.3$_{1.7}$ & 31.3$_{2.1}$ & 31.9$_\text{0.3}$\\

& SR-noC  & 
\textbf{75.0}$_{1.4}$ & 23.0$_{0.8}$ & 35.2$_{1.0}$ & 
67.3$_{0.4}$ & 18.6$_{1.1}$ & 29.1$_{1.4}$ & 
70.2$_{2.5}$ & 20.6$_{1.5}$ & 31.8$_{1.5}$ & 32.0$_\text{0.2}$\\

& \method{}  & 
73.5$_{2.3}$ & \textbf{24.0}$_{0.8}$ & \textbf{36.2}$_{1.2}$ & 
68.4$_{1.3}$ & 17.7$_{0.1}$ & 28.1$_{0.2}$ & 
69.8$_{1.2}$ & \textbf{22.1}$_{1.2}$ & \textbf{33.5}$_{1.2}$ & \textbf{32.6}$_\text{0.9}$\\ \midrule

\multirow{5}{*}{\textbf{Phi}} & 
ZS  & 
64.2$_{1.2}$ & 9.8$_{1.1}$ & 17.0$_{1.6}$ & 
51.2$_{1.0}$ & 14.8$_{0.9}$ & 23.0$_{0.9}$ & 
54.4$_{1.4}$ & 19.3$_{4.4}$ & 28.3$_{4.7}$ & 22.7$_\text{2.0}$\\

& CoT  & 
59.9$_{0.9}$ & 11.7$_{1.1}$ & 19.5$_{1.5}$ & 
\textbf{57.4}$_{1.5}$ & 8.5$_{1.3}$ & 14.8$_{2.1}$ & 
56.2$_{2.1}$ & 11.7$_{2.4}$ & 19.3$_{3.2}$ & 17.9$_\text{0.9}$\\

& SR  & 
56.8$_{0.3}$ & 15.1$_{3.9}$ & 23.7$_{4.9}$ & 
49.1$_{0.6}$ & 13.0$_{0.2}$ & 20.6$_{0.2}$ & 
52.5$_{1.4}$ & 17.2$_{3.5}$ & 25.8$_{4.2}$ & 23.4$_\text{2.5}$\\
 
& SR-noC  & 
59.5$_{2.5}$ & 11.8$_{3.0}$ & 19.6$_{4.3}$ & 
51.2$_{3.7}$ & \textbf{15.1}$_{2.2}$ & 23.3$_{2.6}$ & 
54.1$_{2.1}$ & 17.4$_{4.2}$ & 26.1$_{4.9}$ & 23.0$_\text{1.2}$\\

& \method{}  & 
\textbf{65.2}$_{5.2}$ & \textbf{15.6}$_{2.3}$ & \textbf{25.2}$_{3.0}$ & 
55.8$_{2.1}$ & 15.0$_{0.4}$ & \textbf{23.6}$_{0.4}$ & 
\textbf{57.6}$_{1.3}$ & \textbf{20.0}$_{1.0}$ & \textbf{29.7}$_{1.1}$ & \textbf{26.2}$_\text{1.5}$\\ \bottomrule
\end{tabular}
\caption{Precision, Recall, and F1 scores of models and baselines on three multimodal humor benchmarks.}
\label{tab:overall}
\end{table*}

\begin{table}[h]
    \small
    \centering
    \begin{tabular}{ll|ccc}
    \toprule
   % &   & \multicolumn{3}{c}{\textbf{MemeCap}} &  \multicolumn{3}{|c|}{\textbf{NewYorker}} &  \multicolumn{3}{c}{\textbf{YesBut}}  \\  
\textbf{Model} & \textbf{Input}  & \textbf{MC} & \textbf{NY} & \textbf{YB} \\ \midrule
% &  \textbf{P} & \textbf{R} & \textbf{F1} & \textbf{P} & \textbf{R} & \textbf{F1} \\ \midrule
\multirow{3}{*}{\textbf{GPT4o}} & 
% desc  & 
% \textbf{81.5}$_{2.9}$ & 34.4$_{3.0}$ & 48.3$_{2.7}$ & 
% 79.4$_{1.0}$ & 41.1$_{2.0}$ & 54.1$_{1.6}$ & 
% 80.2$_{2.9}$ & 46.4$_{4.4}$ & 58.7$_{4.1}$\\ 
% & 
Imp  & 
 47.6$_{1.3}$ & 
 53.3$_{0.3}$ & 
 54.9$_{5.3}$\\ 
 
 & Cand  & 
 50.0$_{1.9}$ & 
 56.5$_{2.7}$ & 
 59.7$_{3.1}$\\ 
 
 &  \textbf{Ours}  & 
 \textbf{51.5}$_{0.3}$ & 
 \textbf{58.2}$_{0.5}$ & 
 \textbf{60.5}$_{2.5}$ \\ \midrule
 
\multirow{3}{*}{\textbf{Flash1.5}} & 
% desc  & 
% \textbf{82.1}$_{0.9}$ & \textbf{21.7}$_{2.3}$ & \textbf{34.3}$_{3.0}$ & 
% 77.6$_{2.5}$ & 20.5$_{1.1}$ & 32.4$_{1.5}$ & 
% 80.7$_{1.6}$ & 28.7$_{2.6}$ & 42.4$_{3.1}$\\ 
% & 
Imp  & 
 32.5$_{0.8}$ & 
 36.8$_{0.2}$ & 
 39.0$_{5.1}$\\ 
 
 & Cand  & 
 32.8$_{3.2}$ & 
 \textbf{37.7}$_{1.1}$ & 
 \textbf{43.7}$_{2.7}$\\ 
 
 & \textbf{Ours}  & 
 \textbf{32.9}$_{2.2}$ & 
 36.7$_{1.3}$ & 
 43.1$_{4.7}$\\ \midrule
 
\multirow{3}{*}{\textbf{Qwen2}} & 
% desc  & 
% 73.9$_{1.7}$ & 24.0$_{1.4}$ & 36.2$_{1.5}$ & 
% 76.0$_{2.3}$ & 17.6$_{1.3}$ & 28.6$_{1.8}$ & 
% 74.6$_{1.3}$ & 22.6$_{2.0}$ & 34.6$_{2.3}$\\ 
% & 
Imp  & 
36.2$_{2.1}$ & 
 \textbf{29.2}$_{0.9}$ & 
 \textbf{36.2}$_{1.2}$\\ 
 
 & Cand  & 
 \textbf{37.0}$_{1.0}$ & 
 \textbf{29.2}$_{0.9}$ & 
 33.5$_{0.7}$\\ 
 
 & \textbf{Ours}  & 
 36.2$_{1.2}$ & 
 28.1$_{0.2}$ & 
 33.5$_{1.2}$\\ \midrule
 
\multirow{3}{*}{\textbf{Phi}} & 
% desc  & 
% 69.6$_{1.0}$ & 16.2$_{2.7}$ & 26.2$_{3.5}$ & 
% 64.2$_{1.2}$ & 12.4$_{1.1}$ & 20.7$_{1.6}$ & 
% 67.3$_{1.4}$ & 17.4$_{1.2}$ & 27.7$_{1.6}$\\ 
% &
 Imp  & 
 23.2$_{4.0}$ & 
 23.2$_{1.4}$ & 
 26.2$_{4.2}$\\ 
 
 & Cand  & 
 \textbf{27.3}$_{2.2}$ & 
 23.1$_{0.8}$ & 
 28.1$_{4.9}$\\ 
 
 & \textbf{Ours}  & 
 25.2$_{3.0}$ & 
 \textbf{23.6}$_{0.4}$ & 
 \textbf{29.7}$_{1.1}$\\ \bottomrule

\end{tabular}
\caption{F1 score comparison of using a single refined input: implications (Imp) or candidate explanations (Cand) vs. using both.}
\vspace{-10pt}
\label{tab:ablation}
\end{table}

\begin{figure}[t]
  \includegraphics[width=\linewidth]{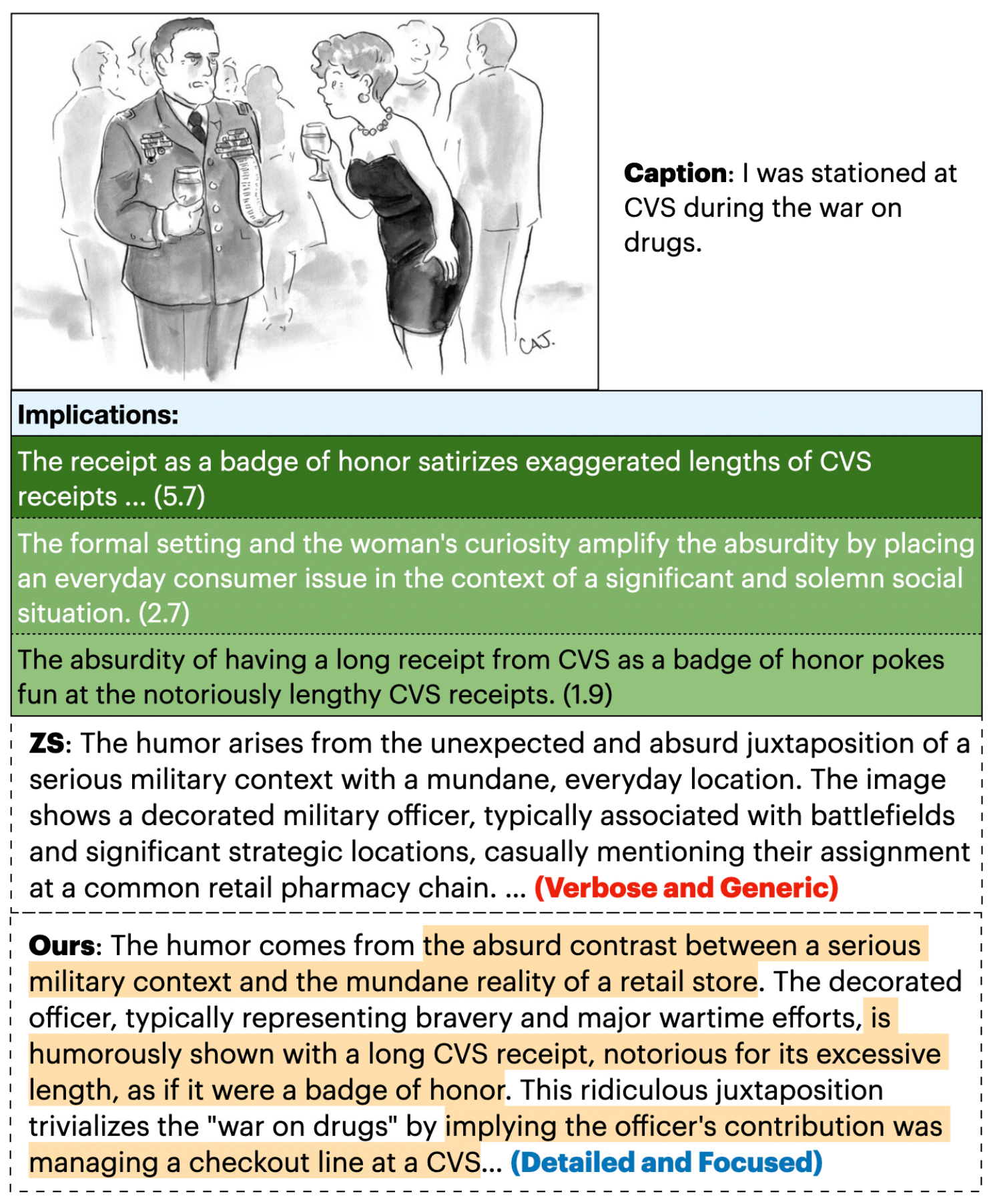} \hfill
  % \vspace{-10pt}
  \caption{An example analysis of the explanations of \base{} and \method{} for a New Yorker Cartoon, using SentenceSHAP. Implications are sorted according to their SentenceSHAP score from most to least important.}
  \label{fig:shap-example}
\end{figure}

We present the comparison of \method{} to the baselines (\S\ref{sec:results:rq1}), look into the contribution of each individual component in our method (\S\ref{sec:results:rq2}), justify the IB framework (\S\ref{sec:results:rq3}), and present an error analysis of our method's predictions  (\S\ref{sec:results:rq4}).  

% \begin{itemize}
%   \item RQ1. How does the performance of our method compare to that of VLM Zero-shot or CoT variants? Table \ref{tab:overall}
% % zs: memecap: 3.4, 4.0, 1.4, 8.2: 4.25
% % zs: newyorker: 4.3, 0.1, 0.6 : 1.6
% % zs: yesbut: 2.4, 2.0, 2.8, 1.4: 2.15
% % cot: memecap: 4.0, 6.2, 2.6, 5.7: 4.62
% % cot: newyorker: 19.4, 14.4, 9.1, 8.8 : 12.9
% % cot: yesbut: 21.7, 16.0, 10.8, 10.4: 14.7
% best memecap: 3.4,4.0, 1.4, 8.2
% best newyorker: 4.3, 0.1, 0.6, 1.6
% best yesbut: 2.4, 2.0, 2.8, 1.4
%   \item RQ2. How does our method compare to VLMs that undergo iterative refinement? Table \ref{tab:overall}
%   \item RQ3. Why do we need IB? TBD.. Feel free to suggest something
%   \item RQ4. What types of input provide the most useful information for our method? Table \ref{tab:ablation}
%   \item RQ5. What are the key success and failure cases in our method's application?
% \end{itemize} 

% sr: memecap: 6.7, 2.0, 0.1, 1.5: 2.57
% sr: newyorker: 2.2, 1.8, 3.0: 2.3
% sr: yesbut: 4.6, 1.8, 2.2, 3.9: 3.12
% sr-no: memecap: 4.0, 1.6, 1.0, 5.6: 3.05
% sr-no: newyorker: 1.3, 3.8, 0.3: 1.8
% sr-no: yesbut: 3.9, 4.6, 1.7, 3.6: 3.45

\subsection{Comparison to the Baselines}
\label{sec:results:rq1}
% While Phi performs the worst among the larger models, our method still boosts its performance by an average of 3.5 F$_1$ points across datasets.
Table~\ref{tab:overall} presents the overall experimental results. Compared to the best of \base{} and \chain{}, \method{} improves an average of 4.2, 1.6, and 2.1 $F_1$ points on the MemeCap, NewYorker, and YesBut datasets, respectively, across models. Among all models, GPT-4o performs best, averaging 3.4 F$_1$ point improvement across datasets. 
\method{} significantly boosts recall while maintaining comparable precision. This suggests that our method effectively integrates external knowledge to generate more comprehensive final explanations, with a slight precision drop due to potential noise.

\base{} performs reasonably well, likely due to these strong VLMs trained on similar tasks. However, \chain{} causes a substantial performance drop. We observe that \chain{}'s reasoning often leads the model to produce more generic explanations and lose focus on explaining the humor.

The self-refine baselines perform similarly to \base{}, with \critic{} slightly outperforming \nocritic{}. This suggests that merely refining the output without adding new information might not be beneficial for these tasks. Furthermore, incorrect feedback from \critic{} could even negatively impact the performance.  
In contrast, \method{} outperforms both self-refinement baselines, improving an average of 2.8, 2.0, and 3.3 $F_1$ points on the MemeCap, NewYorker, and YesBut datasets, respectively; supporting our hypothesis that humor understanding requires additional world knowledge, which \method{} can successfully integrate into the reasoning process.
% , highlighting the importance of incorporating world knowledge in humor understanding

\subsection{Contribution of Individual Components} 
\label{sec:results:rq2}

Since our method introduces several modifications to the standard prompting approach, we assess the contribution of each individual component to the final performance. We conduct ablation tests and employ an explainability technique to point to the features that the model relies on most. 

\paragraph{Ablation study.} Table~\ref{tab:ablation} presents an ablation study where only a single input is provided after refining implications and candidate explanations. GPT-4o and Phi perform better with both inputs, suggesting they effectively integrate relevant information from both to generate improved explanations. In contrast, Flash-1.5 and Qwen2 models rely more on the candidate explanations, which contain more readily-useful information than the implications, indicating these models are less proficient at ignoring noisy or irrelevant implications. 
% This indicates that when implications include distracting or irrelevant information, these models are more likely to incorporate it into the final explanation, decreasing the quality of the final explanation.

\paragraph{Feature importance.} To further pinpoint the contribution of individual implications to the final explanations, we turn to interpretability methods.  
We adapt TokenSHAP \cite{horovicz-goldshmidt-2024-tokenshap}, which estimates the importance of individual tokens to the model's prediction using Monte Carlo Shapley value estimation, to a sentence-level variation that we refer to as SentenceSHAP (see Appendix~\ref{app:sentence-shap} for details). This approach visualizes each sentence's contribution to the final explanation, as shown in Figure~\ref{fig:shap-example}. The explanation from \base{} misses the humor in the long CVS receipt that the officer is holding as a badge of honor, while \method{} is directly informed by the top implication. 

\subsection{Assessment of the IB Framework} 
\label{sec:results:rq3}
\begin{figure}[t]
\small
\centering
  \includegraphics[width=0.9\linewidth]{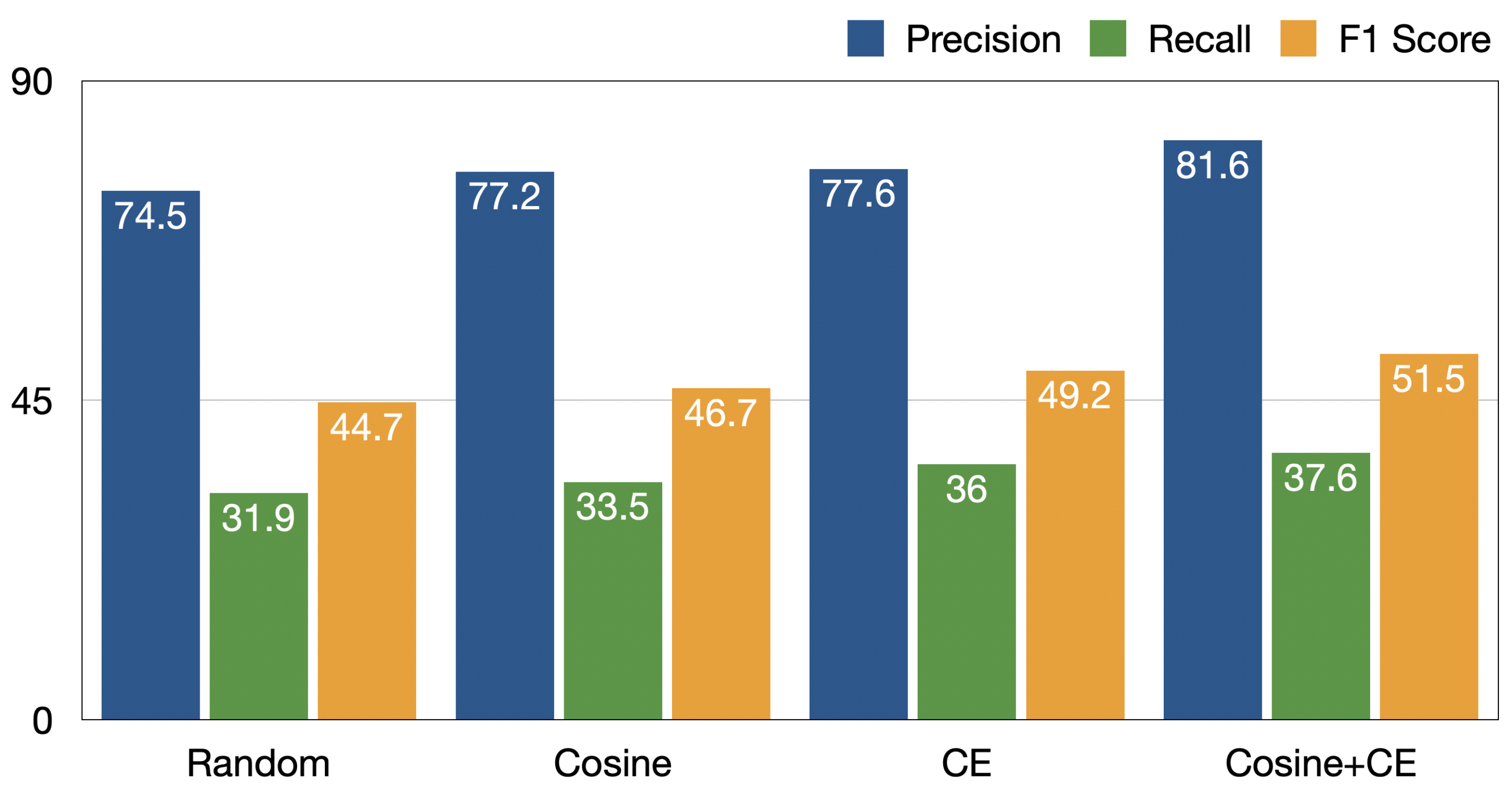}
  \caption {Performance of GPT4o on different IB components.}
  \label{fig:ib-performance}
\end{figure}

\paragraph{IB component analysis.}
We focus on GPT4o, the best performing model across all datasets, and analyze the contribution of each IB component in our method through ablation tests. We evaluate four implication selection approaches (iterative refinement; Sec.~\ref{sec:method:refine-imps}): (1) \textit{Random}, where implications are selected randomly; (2) \textit{Cosine}, which selects implications with the lowest cosine similarity to the previous inputs; (3) \textit{CE}, which selects implications that yield the lowest cross-entropy value when we condition on them to generate the candidate explanations; and (4) \textit{Cosine+CE}, our method presented in Sec.~\ref{sec:method:refine-imps} that combines cosine similarity and cross-entropy based on the IB principle. We conduct the analysis on 100 random instances from each dataset. Figure~\ref{fig:ib-performance} shows that \textit{Cosine+CE} method outperforms the \textit{Cosine} and \textit{CE} baselines, improving $F_1$ score by 4.8 and 2.3 points, respectively, confirming the importance of balancing reducing redundancy with increasing the signal. 

\paragraph{Quality of intermediate explanations.} 
To analyze whether the candidate explanations improve across iterations, we randomly sample 50 examples from each dataset and their outputs generated by GPT-4o and Flash1.5. Since each iteration generates three candidate explanations, we report the highest $F_1$ score among them, and the corresponding precision and recall values in Table~\ref{tab:hop-analysis}. For GPT-4o, $F_1$ scores consistently improve across iterations, primarily driven by recall, which increases by an average of 11.4 points at $h_2$ compared to the initial hop. Precision also improves significantly at $h_1$, averaging an 8.0 point gain across datasets, then stabilizes.
A similar trend is observed in Flash1.5-8B, a considerably smaller model, except for the MemeCap, where $F_1$ scores peak at $h_1$ but decrease by 2.5 points at $h_2$. While precision remains similar at the final hop compared to $h_1$, recall drops by 2.4 points, suggesting smaller models are more susceptible to noisy information as iterations progress.

\begin{table}[h]
\small
\centering
\begin{tabular}{lc|ccc|ccc} % Added one more column (l)
\toprule
& & \multicolumn{3}{c|}{\textbf{GPT4o}} & \multicolumn{3}{c}{\textbf{Flash1.5}} \\
 & \textbf{} & $\mathbf{h_{0}}$ & $\mathbf{h_{1}}$ & $\mathbf{h_{2}}$ & $\mathbf{h_{0}}$ & $\mathbf{h_{1}}$ & $\mathbf{h_{2}}$ \\
\midrule
\textbf{MC} & P & 88.5 & \textbf{92.7} & \textbf{92.7} & 81.0 & 92.2 & \textbf{92.3} \\
  & R & 35.6 & 47.0 & \textbf{48.5} & 21.0 & \textbf{35.0} & 32.6\\
 & $F_1$ & 50.8 & 62.4 & \textbf{63.6} & 33.3 & \textbf{50.7} & 48.2\\
\midrule
\textbf{NY} & P & 79.6 & \textbf{86.5} & 84.7 & 72.2 & \textbf{83.5} & \textbf{83.5}\\
 & R & 50.6 & 57.9 & \textbf{62.8} & 22.9 & 33.4 & \textbf{34.7} \\
 & $F_1$ & 61.9 & 69.4 & \textbf{72.1} & 34.8 & 47.7 & \textbf{49.0} \\
\midrule
\textbf{YB} & P & 67.2 & \textbf{82.3} & 82.0  & 81.0 & 92.2 & \textbf{92.3} \\
 & R & 48.2 & 56.2 & \textbf{57.6} & 26.2 & 36.8 & \textbf{38.1} \\
 & $F_1$ & 56.2 & 66.8 & \textbf{67.6} & 39.6 & 52.6 & \textbf{54.0}\\
\bottomrule
\end{tabular}
\caption{Precision, Recall, and $F_1$ scores on intermediate explanations across hops. h stands for hop. In our experiments, hop $h=0$ corresponds to $k=0$ (no implications), $h=1$ allows up to $k=3$ implications, and $h=2$ allows up to $k=6$ implications.}
\label{tab:hop-analysis}
% \vspace{-10pt}
\end{table}

\paragraph{Error analysis.} 
\label{sec:results:rq4}
We manually analyzed 40 randomly sampled explanations across different models where implications negatively impacted performance. The two most common errors are: dilution of focus (81.2\%) and introducing irrelevant information (18.7\%).
Dilution of focus occurs when implications repeat the same concept multiple times or include overly generalized statements that override more specific details. Irrelevant information, such as common phrases unrelated to the humor can also distort the explanation. See Appendix \ref{app:error-analysis-shap} for examples analyzed using SentenceSHAP.

\section{Conclusions}
\label{sec:conclusion}
We introduced \method{}, an unsupervised method inspired by the information bottleneck principle that addresses  humor explanation tasks by eliciting relevant knowledge from VLMs and iteratively refining the explanation. Our experiments show that \method{} outperforms a range of baselines on three datasets, underscoring the importance of incorporating relevant world knowledge in humor understanding. Our analysis offers insights into the impact of individual components in our method, and justifies the use of the IB principle. We further propose an LLM-based evaluation framework and an adaptation of an interpretability technique. While we tested our contributions in the context of humor interpretation, future work can adapt them to any task that can benefit from eliciting and reasoning on world knowledge.

\section*{Limitations}
\label{sec:limitations}
\paragraph{Subjective nature of humor understanding.} Individuals may interpret humor differently based on their personal background knowledge. While we find that the reference in the data is likely the most representative interpretation of the humor in the image and caption, other interpretations can also be valid, which are not captured in our scores.

\paragraph{Evaluation of explanations.} Humor explanations are often nuanced and subtle. While breaking down the explanation into atomic sentences helps the model verify the accuracy and relevance of each claim, it may overlook the nuanced meaning that emerges when all the sentences are combined.

\paragraph{Trade-off between interpretability and efficiency.} Our method emphasizes interpretable, step-by-step controllable reasoning for the humor explanation tasks, but this comes with increased resource cost. While the computational cost can be managed by limiting the number of implications or image descriptions, the increased cost remains an inherent trade-off for incorporating interpretable reasoning steps. In contrast, less interpretable or controllable approaches may offer greater efficiency. Each call typically involves $\leq$500 input tokens and $\leq$128 output tokens, with up to 20 calls per sample. For 100 samples, this results in an estimated total cost of up to \$4–5 USD using GPT-4o and up to \$1 USD using Gemini-Flash-1.5-8B. 

\section*{Ethics Statement}
\label{sec:ethics}
\paragraph{Data.}
All datasets used in our work, MemeCap, NewYorker, and YesBut, are publicly available. The datasets include images, accompanying texts, and humor interpretations collected from humans and may contain offensive content to some people.

\paragraph{Models.} 
The LLMs and VLMs we used for the experiments are trained on a large-scale web corpora and some of them utilize human feedback. Given their training sources, they could potentially generate content (i.e., descriptions, implications, and explanations) that exhibit societal biases.

\paragraph{Data Collection.} 
We use CloudResearch to collect judgments about model-generated explanations in order to validate our proposed  automatic evaluation method.  To ensure the quality of evaluation, we required that workers were located in English-speaking countries (e.g. US, UK, Canada, Australia, and New Zealand), and had an acceptance rate of at least 93\% on 1,000 prior annotations. We paid \$0.20 for the evaluation task, which means that annotators were compensated with an average hourly wage of \$13, which is comparable to the US minimum wage. We did not use any personal information from annotators. We obtained ethics approval from our institution's research ethics board prior to running the study. 

% TODO: when we prepare the camera-ready version, we should thank Jack in the acknowledgements
\section*{Acknowledgements}
This work was funded, in part, by the Vector Institute for AI, Canada CIFAR AI Chairs program, Accelerate Foundation Models Research Program Award from Microsoft, an NSERC discovery grant, and a research gift from AI2. We thank Jack Hessel, Benyamin Movassagh, Sahithya Ravi, Aditya Chinchure, and Vasile Negrescu for insightful discussions and feedback.

\bibliography{custom,anthology}

\appendix
\section{Dataset Examples}
\label{app:dataset-eg}
\begin{figure*}[t]
  \includegraphics[width=\linewidth]{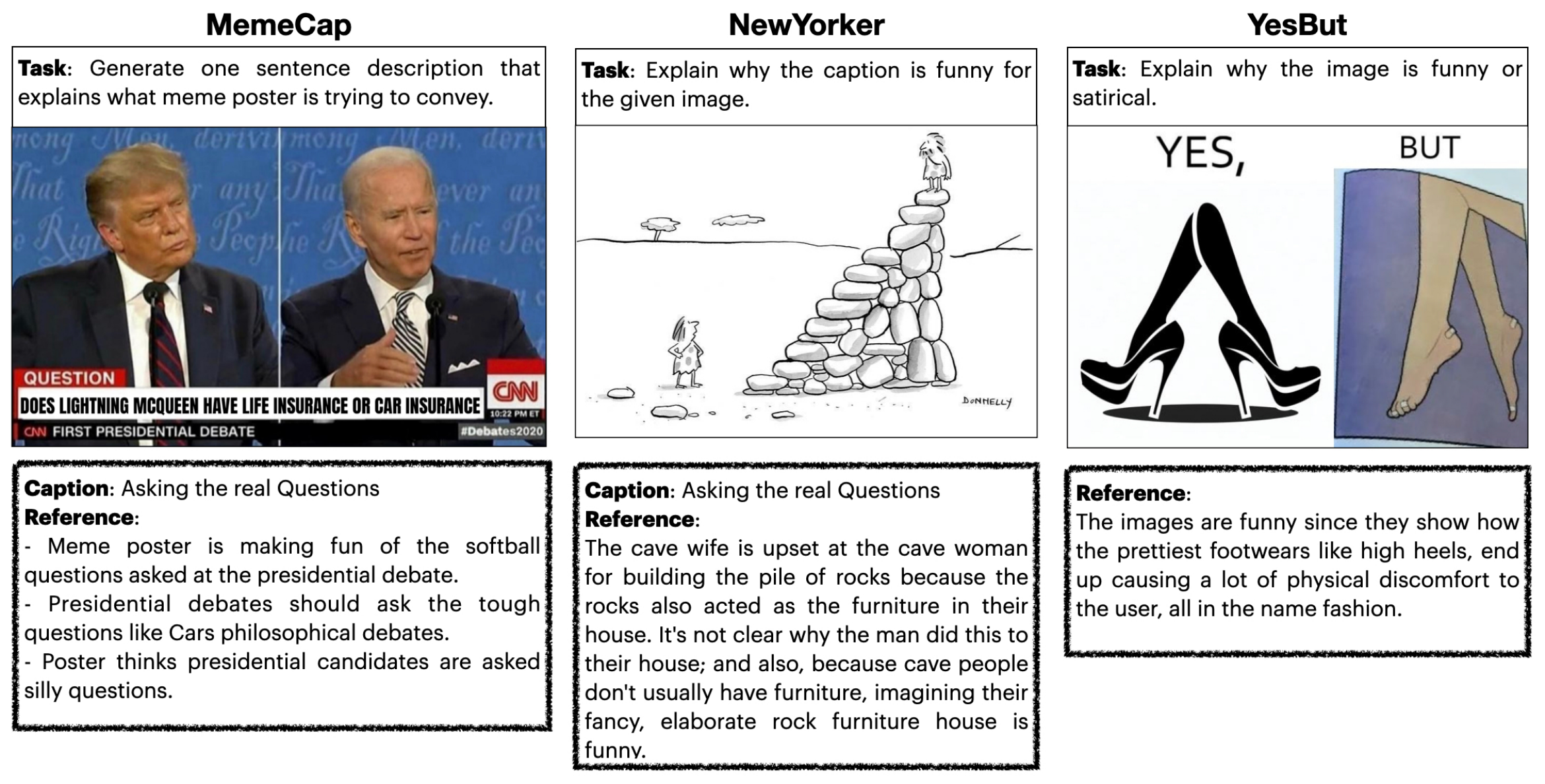} \hfill
  \caption {Dataset Examples on MemeCap, NewYorker, and YesBut.}
  \label{fig:dataset-eg}
\end{figure*}

Figure \ref{fig:dataset-eg} illustrates example data instances from MemeCap, NewYorker, and YesBut.

\section{SentenceSHAP}
\label{app:sentence-shap}
In this section, we introduce SentenceSHAP, an adaptation of TokenSHAP \cite{horovicz-goldshmidt-2024-tokenshap}. While TokenSHAP calculates the importance of individual tokens, SentenceSHAP estimates the importance of individual sentences in the input prompt. The importance score is calculated using Monte Carlo Shapley Estimation, following the same principles as TokenSHAP.

Given an input prompt \( X = \{x_1, x_2, \dots, x_n\} \), where \( x_i \) represents a sentence, we generate all possible combinations of \( X \) by excluding each sentence \( x_i \) (i.e., \( X - \{x_i\} \)). Let \( Z \) represent the set of all combinations where each \( x_i \) is removed. To estimate Shapley values efficiently, we randomly sample from \( Z \) with a specified sampling ratio, resulting in a subset \( Z_s = \{X_1, X_2, \dots, X_s\} \), where each \( X_i = X - \{x_i\} \).

Next, we generate a base response \( r_0 \) using a VLM (or LLM) with the original prompt \( X \), and a set of responses \( R_s = \{r_1, r_2, \dots, r_s\} \), each generated by a prompt from one of the sampled combinations in \( Z_s \).

We then compute the cosine similarity between the base response \( r_0 \) and each response in \( R_s \) using Sentence Transformer (\texttt{BAAI/bge-large-en-v1.5}). The average similarity between combinations with and without \( x_i \) is computed, and the difference between these averages gives the Shapley value for sentence \( x_i \). This is expressed as:
\begin{align}
\notag
\phi(x_i) = \\ \notag
&\frac{1}{s} \sum_{j=1}^{s} \left( \text{cos}(r_0, r_j \mid x_i) - \text{cos}(r_0, r_j \mid \neg x_i) \right)
\end{align}
where \( \phi(x_i) \) represents the Shapley value for sentence \( x_i \), $\text{cos}(r_0, r_j \mid x_i)$ is the cosine similarity between the base response and the response that includes sentence $x_i$, $\text{cos}(r_0, r_j \mid \neg x_i)$ is the cosine similarity between the base response and the response that excludes sentence $x_i$, and $s$ is the number of sampled combinations in $Z_s$.

\section{Error Analysis Based on SentenceSHAP}
\begin{figure*}[t]
  \includegraphics[width=\linewidth]{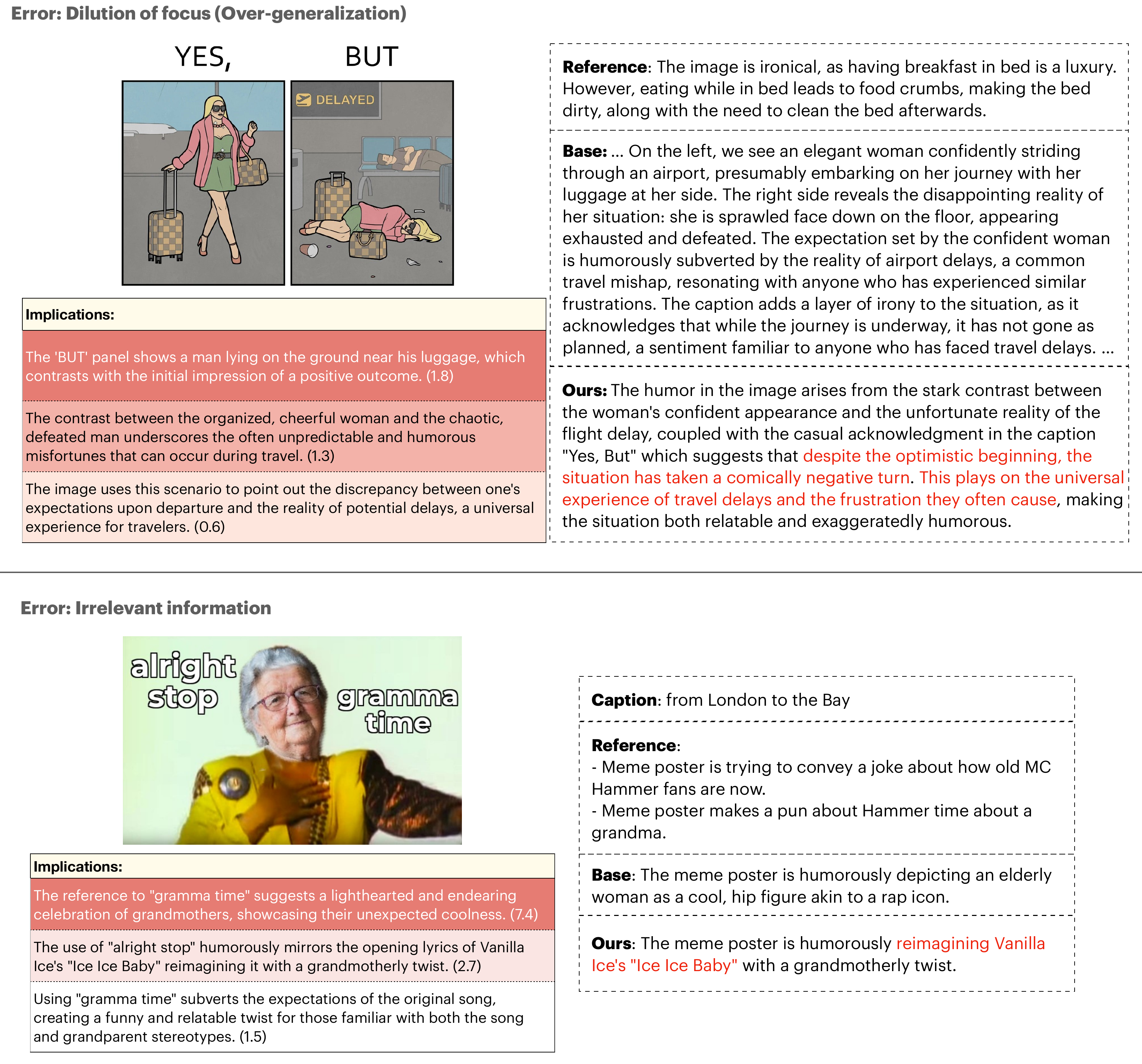} \hfill
  \caption {Examples of negative impact from implications from Phi (top) and GPT4o (bottom).}
  \label{fig:error-analysis}
\end{figure*}

Figure \ref{fig:error-analysis} presents two examples of negative impacts from implications: dilution of focus and the introduction of irrelevant information.
\label{app:error-analysis-shap}

\section{Details on human anntations}
\label{app:cloudresearch}
\begin{figure*}[t]
  \includegraphics[width=\linewidth]{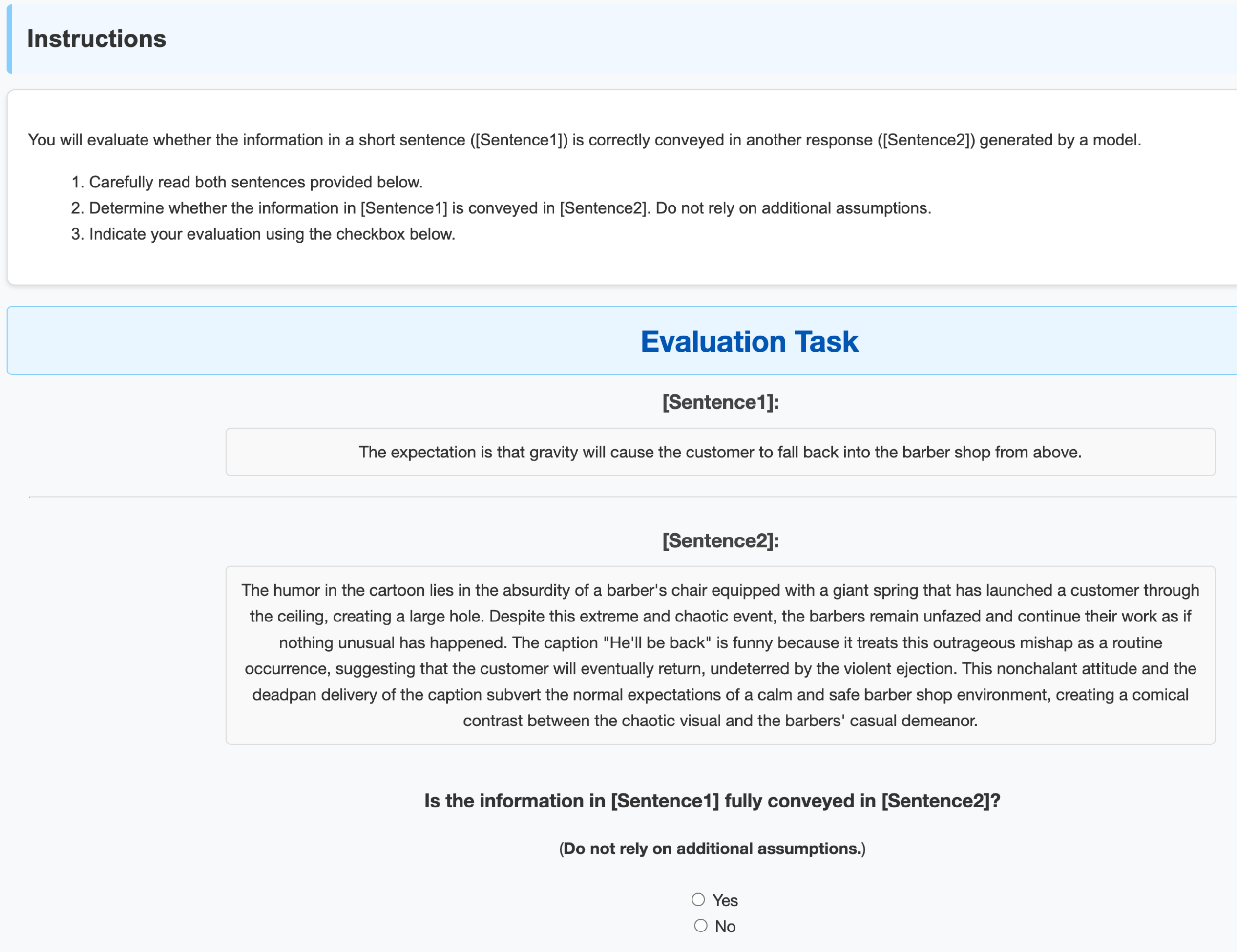} \hfill
  \caption {Annotation interface on CloudResearch used for human evaluation to validate our evaluation metric.}
  \label{fig:cloud-research}
\end{figure*}

We present the annotation interface on CloudResearch used for human evaluation to validate our evaluation metric in Figure \ref{fig:cloud-research}. Refer to Sec.~\ref{sec:ethics} for details on annotator selection criteria and compensation.

\section{Effect of $\alpha$ on explanation quality.}
\label{app:alpha-effect}
To understand the role of the hyperparameter $\alpha$ in Eq.~(1), we conducted an ablation study evaluating F$_1$ scores from GPT-4o across all three datasets (100 samples each), using a fixed random seed. As shown in Table~\ref{tab:alpha-ablation}, performance varies with different $\alpha$ values. When $\alpha=0.0$, the model prioritizes relevance alone; when $\alpha=1.0$, it focuses exclusively on compression. Across datasets, the best performance tends to occur near $\alpha=0.7$, indicating that a balanced trade-off between compression and relevance yields the most informative and accurate intermediate explanations.

\begin{table}[t]
\centering
\begin{tabular}{lcccc}
\toprule
\textbf{Dataset} & $\boldsymbol{0.0}$ & $\boldsymbol{0.3}$ & $\boldsymbol{0.7}$ & $\boldsymbol{1.0}$ \\
\midrule
\textsc{Memecap}     & 46.7 & 51.5 & 51.5 & 48.2 \\
\textsc{NewYorker}   & 57.5 & 57.6 & 57.7 & 57.1 \\
\textsc{YesBut}      & 58.1 & 56.2 & 59.0 & 55.0 \\
\bottomrule
\end{tabular}
\caption{Ablation study for the hyperparameter $\alpha$ used in Eq.~(1).}
\label{tab:alpha-ablation}
\end{table}

\section{Generation Prompts for Selection and Refinement}
\label{app:gen-prompts}
Figures \ref{fig:desc-prompt}, \ref{fig:seed-imp-prompt}, and \ref{fig:nonseed-imp-prompt} show the prompts used for generating image descriptions, seed implications (1st hop), and non-seed implications (2nd hop onward). Figure \ref{fig:cand-prompt} displays the prompt used to generate candidate and final explanations. Image descriptions are used for candidate explanations when existing data is insufficient but are not used for final explanations. For calculating Cross Entropy values (used as a relevance term), we use the prompt in Figure \ref{fig:cand-prompt}, substituting the image with image descriptions, as LLM is used to calculate the cross entropies.

\begin{figure*}[h]
\small
\begin{tcolorbox}[
    title=Prompt for Image Descriptions,
    colback=white,
    colframe=CadetBlue,
    arc=0pt,        % Remove rounded corners
    outer arc=0pt   % Remove outer rounded corners (important for some styles)    
]

Describe the image by focusing on the noun phrases that highlight the actions, expressions, and interactions of the main visible objects, facial expressions, and people.\\
\\
Here are some guidelines when generating image descriptions:\\
* Provide specific and detailed references to the objects, their actions, and expressions. Avoid using pronouns in the description.\\
* Do not include trivial details such as artist signatures, autographs, copyright marks, or any unrelated background information.\\
* Focus only on elements that directly contribute to the meaning, context, or main action of the scene.\\
* If you are unsure about any object, action, or expression, do not make guesses or generate made-up elements.\\
* Write each sentence on a new line.\\
* Limit the description to a maximum of 5 sentences, with each focusing on a distinct and relevant aspect that directly contribute to the meaning, context, or main action of the scene.\\
\\
Here are some examples of desired output:
---\\
\text{[Description]} (example of newyorker cartoon image):\\
Through a window, two women with surprised expressions gaze at a snowman with human arms.\\
---\\
\text{[Description]} (example of newyorker cartoon image):\\
A man and a woman are in a room with a regular looking bookshelf and regular sized books on the wall.\\
In the middle of the room the man is pointing to text written on a giant open book which covers the entire floor.\\
He is talking while the woman with worried expression watches from the doorway.\\
---\\
\text{[Description]} (example of meme):\\
The left side shows a woman angrily pointing with a distressed expression, yelling ``You said memes would work!''.\\
The right side shows a white cat sitting at a table with a plate of food in front of it, looking indifferent or smug with the text above the cat reads, ``I said good memes would work''.\\
---\\
\text{[Description]} (example of yesbut image):\\
The left side shows a hand holding a blue plane ticket marked with a price of ``\$50'', featuring an airplane icon and a barcode, indicating it's a flight ticket.\\
The right side shows a hand holding a smartphone displaying a taxi app, showing a route map labeled ``Airport'' and a price of ``\$65''.\\
---\\

Proceed to generate the description.\\
\text{[Description]}:

\end{tcolorbox}
\caption{A prompt used to generate image descriptions.} % Add a caption to the figure
\label{fig:desc-prompt}
\end{figure*}

%%%%%%%%%%%%%%%%%%%%%%%%%%% Prompt for implications %%%%%%%%%%%%%%%%%%%%%%%%%%%
\begin{figure*}[t]
\small
\begin{tcolorbox}[
    title=Prompt for Seed Implications,
    colback=white,
    colframe=Green,
    arc=0pt,        % Remove rounded corners
    outer arc=0pt,  % Remove outer rounded corners (important for some styles)    
    % breakable,
]

You are provided with the following inputs:\\
- \text{[}Image\text{]}: An image (e.g. meme, new yorker cartoon, yes-but image)\\
- \text{[}Caption\text{]}: A caption written by a human.\\
- \text{[}Descriptions\text{]}: Literal descriptions that detail the image.\\
\\
\#\#\# Your Task:\\
\texttt{[ One-sentence description of the ultimate goal of your task. Customize based on the task. ]}\\
Infer implicit meanings, cultural references, commonsense knowledge, social norms, or contrasts that connect the caption to the described objects, concepts, situations, or facial expressions.\\
\\
\#\#\# Guidelines:\\
- If you are unsure about any details in the caption, description, or implication, refer to the original image for clarification.\\
- Identify connections between the objects, actions, or concepts described in the inputs.\\
- Explore possible interpretations, contrasts, or relationships that arise naturally from the scene, while staying grounded in the provided details.\\
- Avoid repeating or rephrasing existing implications. Ensure each new implication introduces fresh insights or perspectives.\\
- Each implication should be concise (one sentence) and avoid being overly generic or vague.\\
- Be specific in making connections, ensuring they align with the details provided in the caption and descriptions.\\
- Generate up to 3 meaningful implications.\\
\\
\#\#\# Example Outputs:\\
\#\#\#\# Example 1 (example of newyorker cartoon image):\\
\text{[}Caption\text{]}: ``This is the most advanced case of Surrealism I've seen.''\\
\text{[}Descriptions\text{]}: A body in three parts is on an exam table in a doctor's office with the body's arms crossed as though annoyed.\\
\text{[}Connections\text{]}:\\
1. The dismembered body is illogical and impossible, much like Surrealist art, which often explores the absurd.\\
2. The body’s angry posture adds a human emotion to an otherwise bizarre scenario, highlighting the strange contrast.\\
\\
\#\#\#\# Example 2 (example of newyorker cartoon image):\\
\text{[}Caption\text{]}: ``He has a summer job as a scarecrow.''\\
\text{[}Descriptions\text{]}: A snowman with human arms stands in a field.\\
\text{[}Connections\text{]}:\\
1. The snowman, an emblem of winter, represents something out of place in a summer setting, much like a scarecrow's seasonal function.\\
2. The human arms on the snowman suggest that the role of a scarecrow is being played by something unexpected and seasonal.\\
\\
\#\#\#\# Example 3 (example of yesbut image):\\
\text{[}Caption\text{]}: ``The left side shows a hand holding a blue plane ticket marked with a price of `\$50'.''\\
\text{[}Descriptions\text{]}: The screen on the right side shows a route map labeled ``Airport'' and a price of `\$65'.\\
\text{[}Connections\text{]}:\\
1. The discrepancy between the ticket price and the taxi fare highlights the often-overlooked costs of travel beyond just booking a flight.\\
2. The image shows the hidden costs of air travel, with the extra fare representing the added complexity of budgeting for transportation.\\
\\
\#\#\#\# Example 4 (example of meme):\\
\text{[}Caption\text{]}: ``You said memes would work!''\\
\text{[}Descriptions\text{]}: A cat smirks with the text ``I said good memes would work.''\\
\text{[}Connections\text{]}:\\
1. The woman's frustration reflects a common tendency to blame concepts (memes) instead of the quality of execution, as implied by the cat’s response.\\
2. The contrast between the angry human and the smug cat highlights how people often misinterpret success as simple, rather than a matter of quality.\\
\\
\#\#\# Now, proceed to generate output:\\
\text{[}Caption\text{]}: \texttt{[ Caption ]}\\
\\
\text{[}Descriptions\text{]}:\\
\texttt{[ Descriptions ]}\\
\\
\text{[}Connections\text{]}:

\end{tcolorbox}
\caption{A prompt used to generate seed implications.} % Add a caption to the figure
\label{fig:seed-imp-prompt}
\end{figure*}

%%%%%%%%%%%%%%%%%%%%%%%%%%% Prompt for nonseed implications %%%%%%%%%%%%%%%%%%%%%%%%%%%
\begin{figure*}[t]
\small
%  \begin{tcolorbox}[
%  width=\textwidth,
%  colback={white},
%  title={Title},
%  colbacktitle={DarkGreen},
%  coltitle=white,
%  colframe={DarkGreen},
%  breakable
% ]
 % \parskip=5pt

\begin{tcolorbox}[
    % breakable,
    title=Prompt for Non-Seed Implications (2nd hop onward),
    colback=white,
    colframe=Green,
    arc=0pt,        % Remove rounded corners
    outer arc=0pt,  % Remove outer rounded corners (important for some styles)    
    % breakable,
]

You are provided with the following inputs:\\
- \text{[}Image\text{]}: An image (e.g. meme, new yorker cartoon, yes-but image)\\
- \text{[}Caption\text{]}: A caption written by a human.\\
- \text{[}Descriptions\text{]}: Literal descriptions that detail the image.\\
- \text{[}Implication\text{]}: A previously generated implication that suggests a possible connection between the objects or concepts in the caption and description.\\
\\
\#\#\# Your Task:\\
\texttt{[ One-sentence description of the ultimate goal of your task. Customize based on the task. ]}\\
Infer implicit meanings across the objects, concepts, situations, or facial expressions found in the caption, description, and implication. Focus on identifying relevant commonsense knowledge, social norms, or underlying connections.\\
\\
\#\#\# Guidelines:\\
- If you are unsure about any details in the caption, description, or implication, refer to the original image for clarification.\\
- Identify potential connections between the objects, actions, or concepts described in the inputs.\\
- Explore interpretations, contrasts, or relationships that naturally arise from the scene while remaining grounded in the inputs.\\
- Avoid repeating or rephrasing existing implications. Ensure each new implication provides fresh insights or perspectives.\\
- Each implication should be concise (one sentence) and avoid overly generic or vague statements.\\
- Be specific in the connections you make, ensuring they align closely with the details provided.\\
- Generate up to 3 meaningful implications that expand on the implicit meaning of the scene.\\
\\
\#\#\# Example Outputs:\\
\#\#\#\# Example 1 (example of newyorker cartoon image):\\
\text{[}Caption\text{]}: "This is the most advanced case of Surrealism I've seen."\\
\text{[}Descriptions\text{]}: A body in three parts is on an exam table in a doctor's office with the body's arms crossed as though annoyed.\\
\text{[}Implication\text{]}: Surrealism is an art style that emphasizes strange, impossible, or unsettling scenes.\\
\text{[}Connections\text{]}:\\
1. A body in three parts creates an unsettling juxtaposition with the clinical setting, which aligns with Surrealist themes.\\
2. The body’s crossed arms add humor by assigning human emotion to an impossible scenario, reflecting Surrealist absurdity.\\
... \\
\texttt{[ We used sample examples from the prompt for generating seed implications (see Figure \ref{fig:seed-imp-prompt}), following the above format, which includes [Implication]:. ]}
\\
---\\
\\
\#\#\# Proceed to Generate Output:\\
\text{[}Caption\text{]}: \texttt{[ Caption ]}\\
\\
\text{[}Descriptions\text{]}:\\
\texttt{[ Descriptions ]}\\
\\
\text{[}Implication\text{]}:\\
\texttt{[ Implication ]}\\
\\
\text{[}Connections\text{]}:
\end{tcolorbox}
\caption{A prompt used to generate non-seed implications.} % Add a caption to the figure
\label{fig:nonseed-imp-prompt}
\end{figure*}

%%%%%%%%%%%%%%%%%%%%%%%%%%% Prompt for nonseed implications %%%%%%%%%%%%%%%%%%%%%%%%%%%
\begin{figure*}[t]
\small
%  \begin{tcolorbox}[
%  width=\textwidth,
%  colback={white},
%  title={Title},
%  colbacktitle={DarkGreen},
%  coltitle=white,
%  colframe={DarkGreen},
%  breakable
% ]
 % \parskip=5pt

\begin{tcolorbox}[
    % breakable,
    title=Prompt for Candidate and Final Explanations,
    colback=white,
    colframe=RedViolet,
    arc=0pt,        % Remove rounded corners
    outer arc=0pt,  % Remove outer rounded corners (important for some styles)    
    % breakable,
]

You are provided with the following inputs:\\
- **\text{[}Image\text{]}:** A New Yorker cartoon image.\\
- **\text{[}Caption\text{]}:** A caption written by a human to accompany the image.\\
- **\text{[}Image Descriptions\text{]}:** Literal descriptions of the visual elements in the image.\\
- **\text{[}Implications\text{]}:** Possible connections or relationships between objects, concepts, or the caption and the image.\\
- **\text{[}Candidate Answers\text{]}:** Example answers generated in a previous step to provide guidance and context.\\
\\
\#\#\# Your Task:\\
Generate **one concise, specific explanation** that clearly captures why the caption is funny in the context of the image. Your explanation must provide detailed justification and address how the humor arises from the interplay of the caption, image, and associated norms or expectations.\\
\\
\#\#\# Guidelines for Generating Your Explanation:\\
1. **Clarity and Specificity:**  \\
   - Avoid generic or ambiguous phrases.  \\
   - Provide specific details that connect the roles, contexts, or expectations associated with the elements in the image and its caption.  \\
\\
2. **Explain the Humor:**  \\
- Clearly connect the humor to the caption, image, and any cultural, social, or situational norms being subverted or referenced.  \\
- Highlight why the combination of these elements creates an unexpected or amusing contrast.\\
\\
3. **Prioritize Clarity Over Brevity:**  \\
- Justify the humor by explaining all important components clearly and in detail.  \\
- Aim to keep your response concise and under 150 words while ensuring no critical details are omitted.  \\
\\
4. **Use Additional Inputs Effectively:**\\
- **\text{[}Image Descriptions\text{]}:** Provide a foundation for understanding the visual elements."   \\
- **\text{[}Implications\text{]}:** Assist in understanding relationships and connections but do not allow them to dominate or significantly alter the central idea.\\
- **\text{[}Candidate Answers\text{]}:** Adapt your reasoning by leveraging strengths or improving upon weaknesses in the candidate answers.\\
\\
Now, proceed to generate your response based on the provided inputs.\\
\\
\#\#\# Inputs:\\
\text{[}Caption\text{]}: \texttt{\text{[} Caption \text{]}}\\
\\
\text{[}Descriptions\text{]}:\\
\texttt{\text{[} Top-K Implications \text{]}}\\
\\
\text{[}Implications\text{]}:\\
\texttt{\text{[} Top-K Implications \text{]}}\\
\\
\text{[}Candidate Anwers\text{]}:\\
\texttt{\text{[} Top-K Candidate Explanations \text{]}}\\
\\
\text{[}Output\text{]}:\\

\end{tcolorbox}
\caption{A prompt used to generate candidate and final explanations.} % Add a caption to the figure
\label{fig:cand-prompt}
\end{figure*}

\section{Evaluation Prompts}
\label{app:eval-prompts}
Figures \ref{fig:recall-prompt} and \ref{fig:precision-prompt} present the prompts used to calculate recall and precision scores in our LLM-based evaluation, respectively.

%%%%%%%%%%%%%%%%%%%%%%%%%%% Prompt for nonseed implications %%%%%%%%%%%%%%%%%%%%%%%%%%%
\begin{figure*}[t]
\small
\begin{tcolorbox}[
    % breakable,
    title=Prompt for Evaluating Recall Score,
    colback=white,
    colframe=MidnightBlue,
    arc=0pt,        % Remove rounded corners
    outer arc=0pt,  % Remove outer rounded corners (important for some styles)    
    % breakable,
]

Your task is to assess whether \text{[}Sentence1\text{]} is conveyed in \text{[}Sentence2\text{]}. \text{[}Sentence2\text{]} may consist of multiple sentences.\\
\\
Here are the evaluation guidelines:\\
1. Mark 'Yes' if \text{[}Sentence1\text{]} is conveyed in \text{[}Sentence2\text{]}.\\
2. Mark 'No' if \text{[}Sentence2\text{]} does not convey the information in \text{[}Sentence1\text{]}.\\
\\
Proceed to evaluate. \\
\\
\text{[}Sentence1\text{]}: \texttt{[ One Atomic Sentence from Decomposed Reference Explanation ]} \\
\\
\text{[}Sentence2\text{]}: \texttt{[ Predicted Explanation ]}\\
\\
\text{[}Output\text{]}:

\end{tcolorbox}
\caption{Prompt for evaluating recall score.} % Add a caption to the figure
\label{fig:recall-prompt}
\end{figure*}

\begin{figure*}[t]
\small
\begin{tcolorbox}[
    % breakable,
    title=Prompt for Evaluating Precision Score,
    colback=white,
    colframe=MidnightBlue,
    arc=0pt,        % Remove rounded corners
    outer arc=0pt,  % Remove outer rounded corners (important for some styles)    
    % breakable,
]

Your task is to assess whether \text{[}Sentence1\text{]} is inferable from \text{[}Sentence2\text{]}. \text{[}Sentence2\text{]} may consist of multiple sentences.\\
\\
Here are the evaluation guidelines:\\
1. Mark "Yes" if \text{[}Sentence1\text{]} can be inferred from \text{[}Sentence2\text{]} — whether explicitly stated, implicitly conveyed, reworded, or serving as supporting information.\\
2. Mark 'No' if \text{[}Sentence1\text{]} is absent from \text{[}Sentence2\text{]}, cannot be inferred, or contradicts it.\\
\\
Proceed to evaluate. \\
\\
\text{[}Sentence1\text{]}: \texttt{[ One Atomic Sentence from Decomposed Predicted Explanation ]}\\
\\
\text{[}Sentence2\text{]}: \texttt{[ Reference Explanation ]}\\
\\
\text{[}Output\text{]}:

\end{tcolorbox}
\caption{Prompt for evaluating precision score.} % Add a caption to the figure
\label{fig:precision-prompt}
\end{figure*}

\section{Prompts for Baselines}
\label{app:base-prompts}

Figure \ref{fig:base-prompt} presents the prompt used for the ZS, CoT, and SR Generator methods. While the format remains largely the same, we adjust it based on the baseline being tested (e.g., CoT requires generating intermediate reasoning, so we add extra instructions for that).
Figure \ref{fig:critic-prompt} shows the prompt used in the SR critic model. The critic's criteria include: (1) \textit{correctness}, measuring whether the explanation directly addresses why the caption is humorous in relation to the image and its caption; (2) \textit{soundness}, evaluating whether the explanation provides a well-reasoned interpretation of the humor; (3) \textit{completeness}, ensuring all important aspects in the caption and image contributing to the humor are considered; (4) \textit{faithfulness}, verifying that the explanation is factually consistency with the image and caption; and (5) \textit{clarity}, ensuring the explanation is clear, concise, and free from unnecessary ambiguity.
\begin{figure*}
\small
\begin{tcolorbox}[
    % breakable,
    title=Prompt for Baselines,
    colback=white,
    colframe=Black,
    arc=0pt,        % Remove rounded corners
    outer arc=0pt,  % Remove outer rounded corners (important for some styles)    
    % breakable,
]

You are provided with the following inputs:\\
- **\text{[}Image\text{]}:** A New Yorker cartoon image.\\
- **\text{[}Caption\text{]}:** A caption written by a human to accompany the image.\\
\texttt{[ if Self-Refine with Critic is True: ]} \\
- **\text{[}Feedback for Candidate Answer\text{]}:** Feedback that points out some weakness in the current candidate responses.\\
\texttt{[ if Self-Refine is True: ]} \\
- **\text{[}Candidate Answers\text{]}:** Example answers generated in a previous step to provide guidance and context.\\
\\
\#\#\# Your Task:\\
Generate **one concise, specific explanation** that clearly captures why the caption is funny in the context of the image. Your explanation must provide detailed justification and address how the humor arises from the interplay of the caption, image, and associated norms or expectations.\\
\\
\#\#\# Guidelines for Generating Your Explanation:\\
1. **Clarity and Specificity:**  \\
   - Avoid generic or ambiguous phrases.  \\
   - Provide specific details that connect the roles, contexts, or expectations associated with the elements in the image and its caption.  \\
\\
2. **Explain the Humor:**  \\
- Clearly connect the humor to the caption, image, and any cultural, social, or situational norms being subverted or referenced.  \\
- Highlight why the combination of these elements creates an unexpected or amusing contrast.\\
\\
3. **Prioritize Clarity Over Brevity:**  \\
- Justify the humor by explaining all important components clearly and in detail.  \\
- Aim to keep your response concise and under 150 words while ensuring no critical details are omitted.  \\
\\
\texttt{[ if Self-Refine is True: ]}\\
4. **Use Additional Inputs Effectively:**\\
- **[Candidate Answers]:** Adapt your reasoning by leveraging strengths or improving upon weaknesses in candidate answers. \\
\texttt{[ if Self-Refine with Critic is True: ]}\\
- **[Feedback for Candidate Answer]:** Feedback that points out some weaknesses in the current candidate responses.\\
\\
\texttt{ [ if CoT is True: ]} \\
Begin by analyzing the image and the given context, and explain your reasoning briefly before generating your final response. \\
\\
Here is an example format of the output: \\
\{\{ \\
    "Reasoning": "...", \\
    "Explanation": "..."   \\
\}\} \\

Now, proceed to generate your response based on the provided inputs.\\
\\
\#\#\# Inputs:\\
\text{[}Caption\text{]}: \texttt{\text{[} Caption \text{]}}\\
\\
\text{[}Candidate Answers\text{]}: \texttt{\text{[} Candidate Explanations \text{]}}\\
\\
\text{[}[Feedback for Candidate Answer]:\text{]}: \texttt{\text{[} Feedback for Candidate Explanations \text{]}}\\
\\
\text{[}Output\text{]}:\\

\end{tcolorbox}
\caption{A prompt used for baseline methods, with conditions added based on the specific baseline being experimented with.} % Add a caption to the figure
\label{fig:base-prompt}
\end{figure*}

\begin{figure*}
\small
\begin{tcolorbox}[
    % breakable,
    title=Prompt for Self-Refine Critic,
    colback=white,
    colframe=Black,
    arc=0pt,        % Remove rounded corners
    outer arc=0pt,  % Remove outer rounded corners (important for some styles)    
    % breakable,
]
\texttt{[ Customize goal text here: ]} \\
\texttt{MemeCap:} You will be given a meme along with its caption, and a candidate response that describes what meme poster is trying to convey. \\
\texttt{NewYorker:} You will be given an image along with its caption, and a candidate response that explains why the caption is funny for the given image. \\
\texttt{YesBut:} You will be given an image and a candidate response that describes why the image is funny or satirical. \\
\\
Your task is to criticize the candidate response based on the following evaluation criteria: \\
- Correctness: Does the explanation directly address why the caption is funny, considering both the image and its caption? \\
- Soundness: Does the explanation provide a meaningful and well-reasoned interpretation of the humor? \\
- Completeness: Does the explanation address all relevant aspects of the caption and image (e.g., visual details, text) that contribute to the humor? \\
- Faithfulness: Is the explanation factually consistent with the details in the image and caption? \\
- Clarity: Is the explanation clear, concise, and free from unnecessary ambiguity? \\
 \\
Proceed to criticize the candidate response ideally using less than 5 sentences:\\
\\
\text{[}Caption\text{]}: \texttt{[ caption ]}\\
\\
\text{[}Candidate Response\text{]}: \\
 \texttt{\text{[} Candidate Response \text{]}}\\
\\
\text{[}Output\text{]}: \\
\end{tcolorbox}
\caption{A prompt used in SR critic model.} % Add a caption to the figure
\label{fig:critic-prompt}
\end{figure*}

% \begin{figure*}[t]
%   \includegraphics[width=\linewidth]{figures/error-analysis.pdf} \hfill
%   \vspace{-20pt}
%   \caption {Examples of negative impact from implications from Phi (top) and GPT4o (bottom).}
%   \label{fig:error-analysis}
% \end{figure*}

\end{document}